\documentclass[10pt,twocolumn,letterpaper]{article}

\usepackage{cvpr}              
\definecolor{cvprblue}{rgb}{0.21,0.49,0.74}
\definecolor{myblue}{RGB}{210, 225, 255}
\usepackage[pagebackref,breaklinks,colorlinks,allcolors=cvprblue]{hyperref}
\usepackage{xcolor}         
\usepackage[table]{xcolor} 
\usepackage{multirow}
\usepackage{booktabs}

\def\paperID{***} 
\def\confName{CVPR}
\def\confYear{2026}

\title{OmniGen2: Towards Instruction-Aligned Multimodal Generation}

\author{
Chenyuan Wu$^{1,2}$\thanks{Co-first Authors and Listed in Alphabetical Order}~~,
Jiahao Wang$^{1,3*}$,
Pengfei Zheng$^{1,2*}$,
Ruiran Yan$^{1,2*}$,
Shitao Xiao$^{1*}$\footnotemark[4],
Xin Luo$^{1,2*}$,
\\
Yueze Wang$^{1*}$,
Wanli Li$^{1,4}$\thanks{Core Contributor}~~,
Xiyan Jiang$^{1,4}$\footnotemark[2]~~,
Yexin Liu$^{1}$\footnotemark[2]~~,
Junjie Zhou$^{1}$, 
Ziyi Xia$^{1}$,
\\
Ze Liu$^{1,2}$, 
Chaofan Li$^{1}$, 
Haoge Deng$^{1,3}$, Kun Luo$^{1,3}$, 
Bo Zhang$^{4}$, 
Jiajun Zhang$^{3}$,\\ 
Dong Liu$^{2}$, Defu Lian$^{2}$, Xinlong Wang$^{1}$, Zhongyuan Wang$^{1}$, Tiejun Huang$^{1}$, Zheng Liu$^{1}$\thanks{Corresponding Author}~\thanks{Project Lead}
\\
 $^{1}$ Beijing Academy of Artificial Intelligence, $^{2}$ University of Science and Technology of China, \\
 $^{3}$ Institute of Automation, Chinese Academy of Sciences, $^{4}$ Zhejiang University\\
 \{stxiao, yzwang\}@baai.ac.cn \quad zhengliu1026@gmail.com \\
}

\begin{document}
\maketitle
\begin{abstract}

Multimodal generative models can process instructions in various modalities and demonstrate outstanding performance across a wide range of image generation tasks. However, their robustness in complex real-world scenarios remains limited due to insufficient generalized instruction alignment. We introduce \textbf{OmniGen2}, a unified multimodal generator designed to follow complex, fine-grained instructions. Our core contribution is a two-stage design that first builds a strong, world-knowledge-grounded foundation model and then aligns it using a progressive, multi-task instruction tuning strategy. The foundation model features a streamlined architecture with decoupled decoding for versatile multimodal generation and a novel positional encoding scheme to improve learning efficiency. We ground this model in real-world knowledge using large-scale data construction pipelines. Building on this foundation, we propose a progressive, reinforcement-based alignment process. This phase carefully schedules training tasks and reward signals to foster cross-task knowledge transfer, significantly improving the model's instruction-following capabilities. Our models demonstrate competitive performance on standard benchmarks and our dedicated in-context generation benchmark, \textbf{OmniContext}. We have released our models, code, benchmark, and training datasets at \url{https://github.com/VectorSpaceLab/OmniGen2}.


\end{abstract}    
\section{Introduction}
\label{sec:intro}
\begin{figure*}
    \centering
    \includegraphics[width=0.9\linewidth]{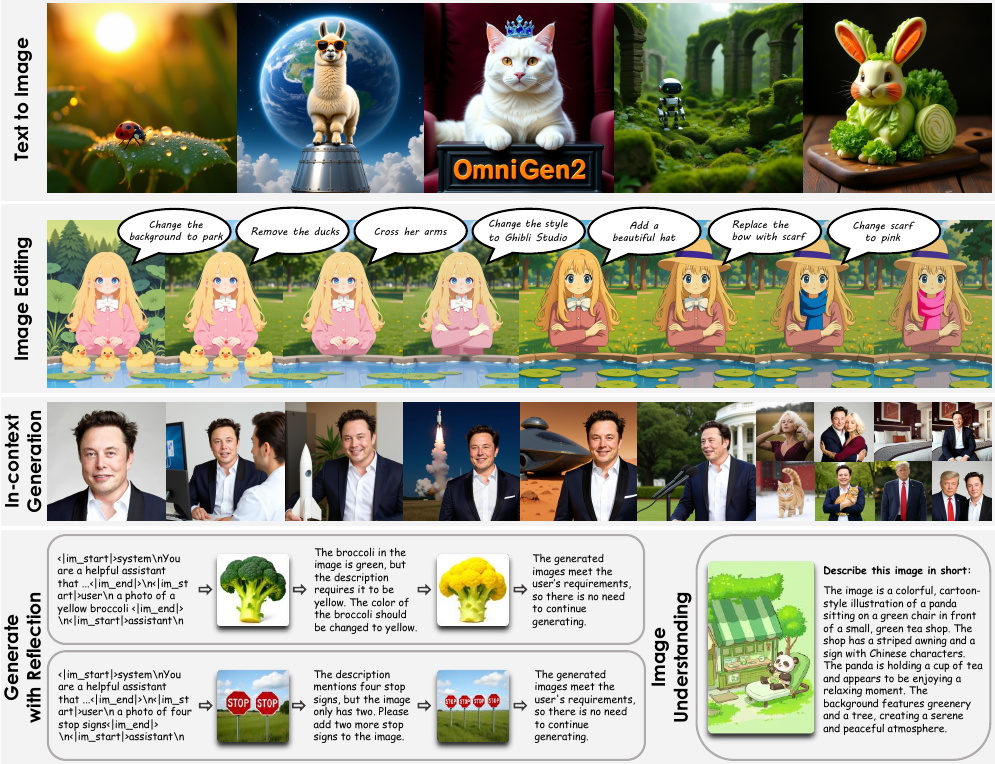}
    \label{fig:overview}
    \caption{Overview of versatile abilities of OmniGen2.}
    \vspace{-10pt}
\end{figure*}

Multimodal image generation has witnessed rapid progress in the past year. Generative models such as GPT-Image-1~\cite{hurst2024gpt}, Flux~\cite{fluxkontext}, Qwen-Image~\cite{wu2025qwenimagetechnicalreport}, Seedream~\cite{seedream2025seedream} and NanoBanana~\cite{google2025gemini25flash} demonstrate increasingly broad and versatile capabilities such as stylization, text rendering, in-context generation, and knowledge-driven generation, marking a significant step toward general-purpose generative intelligence. Given these diverse capabilities, it is essential to perform multimodal instruction alignment to ensure the controllability, semantic consistency and overall generation quality. This involves two key challenges. The first is constructing a robust and versatile foundation model. The model must be endowed with nascent instruction following capabilities and broad world knowledge, while strictly avoiding over-training. The second is aligning the foundation model. This alignment requires explicit and comprehensive reward signals and must ensure consistency across all generation tasks. 

Existing open-source generation models are somewhat deficient as initial base model. Some models are specialized and can not handle tasks beyond their training scope while some are over-optimized towards specific aesthetic preferences, resulting in a severe loss of plasticity. Meanwhile, instruction alignment requires the foundation model to possess a deep understanding of multimodal semantic and task intent. Therefore, we aim to first establish a base model which is simple, versatile and flexible. 

The versatility of a generative model depends a lot on the scale and diversity of its training data. Existing datasets are typically generated either via inpainting models~\cite{wei2024omniedit}, which have limited task coverage, or by retrieving images from the Internet~\cite{xiao2025omnigen}, which results in limited data volume and low image quality. To address this, we develop extensive data construction pipelines that leverage video sources, providing richer in-context and editing examples.

A strong architecture is equally crucial. OmniGen2 achieves unified multimodal generation by conditioning the diffusion transformer on the variable-length hidden states of a Vision Language Model (VLM), effectively leveraging the VLM’s deep semantic understanding and rich world knowledge. To support diverse tasks, we introduce Omni-RoPE, which enhances spatial consistency across images and improves cross-image localization. While conceptually similar to MetaQuery~\citep{pan2025metaquery}, OmniGen2 differs in execution: rather than using fixed-length query tokens, it conditions the diffusion decoder on the VLM’s variable-length hidden states, avoiding information bottlenecks. During the majority of the training process, the VLM is frozen, and optimization focuses on image rendering, making OmniGen2 more efficient than models like Mogao~\citep{liao2025mogao} and BAGEL~\citep{bagel}.

Once the foundation model is obtained via pre-training and fine-tuning, we apply progressive reinforcement learning to facilitate instruction alignment of OmniGen2. Specifically, we adopt Group Relative Policy Optimization~(GRPO)~\cite{shao2024deepseekmath} and divide the instruction alignment process into multiple sequential stages. At each stage, appropriate reward is selected to optimize the alignment for specific target tasks. The training sequence is carefully organized to promote inter-task transfer.

Our extensive evaluation of OmniGen2 reveals its competitive performance across various task domains, including text-to-image (T2I) generation, image editing (Edit), and in-context generation (IC). Instruction alignment consistently and significantly improves the performance of the base model across all these tasks. Notably, for the in-context generation task, there is currently a lack of well-established public benchmark to systematically assess and compare the key capabilities of different models. To mitigate this limitation, we introduce the \textbf{OmniContext} benchmark, comprising eight task categories specifically designed to evaluate consistency across individuals, objects, and scenes.

In summary, our main contributions are as follows:
\begin{itemize}
    \item We introduce OmniGen2, a powerful multimodal generative model that is systematically instruction aligned. The model demonstrates superior instruction following ability, context consistency, and generation quality across diverse task scenarios.
    \item We establish an end-to-end pipeline to achieve comprehensive instruction alignment. This pipeline spans from strong foundation model construction to dedicated multi-task alignment.
    \item We present the OmniContext benchmark, a rigorous suite designed to evaluate in-context image generation, providing the community with a standardized tool to measure progress in this key area.
\end{itemize}

\begin{figure*}[ht!]
    \centering
    \includegraphics[width=1\linewidth]{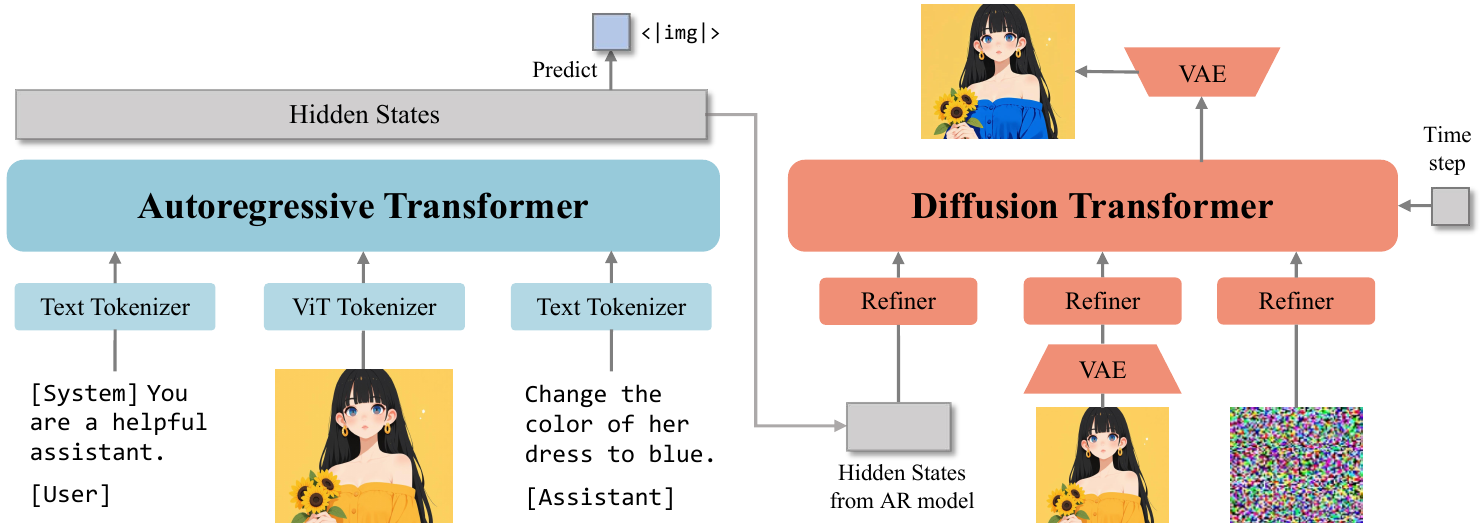}
    \caption{Architecture of OmniGen2. OmniGen2 employs separate transformers for autoregressive and diffusion. Two distinct image encoders are utilized: ViT encodes images for input into the text transformer, while VAE encodes images for the diffusion transformer.}
    \label{fig:model}
    \vspace{-10pt}
\end{figure*}

\section{Dataset Construction}
To build a versatile and robust base model, high-quality, large-scale, and diverse training data is essential. Our goal is to equip the base model with broad world knowledge and the ability to generate varied content. A key challenge is the lack of high-quality public data for complex tasks like detailed image editing and consistent in-context generation. Therefore, we not only gather existing datasets but also build our own scalable pipelines to create the high-quality data needed to fill these gaps.

\noindent\textbf{Foundational Knowledge and General Capabilities.}
To build a strong foundation, we first curate a massive dataset covering both multimodal understanding and text-to-image (T2I) generation. For the former, we adopt LLaVA-OneVision~\cite{li2024llavaoneversion}. For T2I, we collect approximately 140M open-source image-text pairs from diverse datasets~\cite{li2024recaption,chen2023pixart,chen2023sharegpt4v,schuhmann2022laion,chen2024allava,onoe2024docci,li2024DenseFusion,sun2024journeydb,chen2025blip3}, supplemented with 10M proprietary images annotated by Qwen2.5-VL-72B~\cite{bai2025qwen25vl}.

\noindent\textbf{Advanced Capabilities for Editing and In-Context Tasks.}
To address the data scarcity in more complex domains, we develop dedicated construction pipelines. For image editing, we integrate public datasets such as SEED-Data-Edit~\cite{ge2024seededit} and OmniEdit~\cite{wei2024omniedit}, and further construct high-quality editing data using inpainting and video-based pipelines. For in-context generation and editing, we build our datasets from video sources to model consistent subjects across varying scenarios. We employ vision-language models for subject detection, segmentation, and semantic filtering, resulting in diverse and semantically consistent triplets for training.

\noindent\textbf{Fostering Higher-Level Reasoning.}
Finally, to push the base model's capabilities beyond simple generation, we construct interleaved and reflection datasets to enhance temporal reasoning and self-correction capabilities in multimodal models. Detailed pipeline steps, examples and the capability of reflection are provided in Appendix 9.2, 9.3, 9.4, 9.5, 9.6.



\section{Method}

\begin{figure*}[t]
    \setlength{\abovecaptionskip}{3pt}
    \centering
    \includegraphics[width=\linewidth]{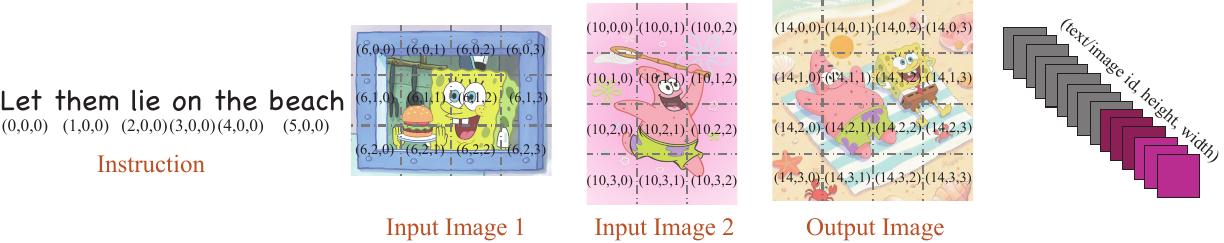}
    \caption{Illustration of \textbf{Omni-RoPE}. Each token in the $k$-th image is assigned a three-dimensional positional identifier $(\Delta_I^{(k)}, h, w)$, where $\Delta_I^{(k)}$ denotes the \textit{instance identity} shared by all tokens within the same image, and $(h,w)$ are the local 2D spatial coordinates computed from $(0,0)$. This decomposition enables the model to distinguish different images while preserving local spatial consistency for tasks such as image editing.}
    \label{fig:omni_rope}
    \vspace{-10pt}
\end{figure*}

OmniGen2 is built on three key components: (1) a decoupled architecture for unified generation, (2) Omni-RoPE for efficient contextual learning, and (3) a multi-stage training and alignment curriculum that progresses from broad knowledge to fine-grained instruction following.


\subsection{Overall Architecture}
\label{sec:architecture}
We aim to design a simple, efficient, and versatile architecture for multimodal generation. Following this principle, OmniGen2 utilizes decoupled pathways for text and image generation. It employs two distinct transformer modules to efficiently facilitate the concurrent support of both understanding and generation capabilities, as illustrated in Figure~\ref{fig:model}. The autoregressive transformer model is initialized from a VLM~(Qwen2.5-VL-3B~\cite{bai2025qwen25vl}). This VLM provides extensive world knowledge and deep understanding of multimodal instructions. The diffusion transformer is randomly initialized and dedicated solely to high-fidelity image synthesis.

The two modules operate in sequence. First, the VLM processes the input multimodal context. A special token, \texttt{<|img|>}, is learned to distinguish the understanding and generation tasks. The generation of this token triggers the image generation. The corresponding hidden states from the VLM are extracted and fed to the diffusion decoder as a condition. It encodes high-level semantic instruction. Besides the high-level semantic encoding from the VLM, we incorporate low-level image features to ensure consistency of fine-grained visual details for tasks like image editing. We utilize Flux-VAE~\citep{FLUX} for this purpose. This approach avoids complex architectural modifications to the pre-trained VLM, thereby preserving its powerful instruction understanding capabilities.



\subsection{Diffusion Decoder}
OmniGen2 employs computational efficient conditioning mechanism for the diffusion decoder. Existing methods such as MetaQuery~\cite{pan2025metaquery} compressing instructions into a fixed set of learnable query tokens can create an information bottleneck. In contrast, OmniGen2 directly leverages the rich hidden states from the VLM's final layer. Furthermore, we utilize only the hidden states corresponding to text tokens, as the VAE features already provide sufficient visual detail.

Within the diffusion decoder, we adopt a unified transformer backbone, following the architecture of Lumina-Image 2.0~\citep{qin2025lumina-image2}, where the parameters are shared across modalities. This design choice is motivated by the motivation that language and vision share substantial semantic representations. Consequently, parameter sharing provides a more natural and efficient means of cross-modal alignment than maintaining separate pathways~\citep{FLUX, fluxkontext,bagel}. Meanwhile, this design facilitates more consistent information exchange between modalities. Before processed by the core transformer blocks, input conditioning signals~(VLM hidden states, VAE features, and noisy latents) are aligned by a lightweight two-layer transformer refiner. This refiner shares the same architecture as the transformer block employed in Lumina-Image 2.0.

\begin{table}[t]
    \centering
    \resizebox{0.99\linewidth}{!}{
        \begin{tabular}{@{}l c c c@{}}
            \toprule
            \textbf{Method} & \textbf{PosID} $\operatorname{PosID}_k(h,w)$ & \textbf{Steps to Target $\downarrow$} & \textbf{Final Loss $\downarrow$} \\
            \midrule
            Lumina-Image-2.0's & $(0,\, h + \Delta_h,\, w + \Delta_w)$ & $\sim$2{,}500 & 0.017 \\
            Qwen2-VL's & $(\Delta_I,\, h + \Delta_I,\, w + \Delta_I)$ & $\sim$1{,}200 & 0.005 \\
            Omni-RoPE (Ours) & \multirow{2}{*}{$(\Delta_I,\, h,\, w)$} & \textbf{$\sim$800} & 0.003 \\
            \quad + Image Index Emb. & & \textbf{$\sim$800} & \textbf{0.002} \\
            \bottomrule
        \end{tabular}
    }
    \caption{
        Comparison of RoPE designs in the toy reconstruction task. 
        Models are trained to reproduce the $k$-th image among randomly sampled inputs. 
        We report the number of steps required to reach the target (\textit{loss} $<$ 0.014). 
        Omni-RoPE achieves both faster convergence and lower final loss. 
        \textbf{Note:} $\Delta_h$ and $\Delta_w$ denote the accumulated coordinate offsets in the height and width dimensions, respectively, while $\Delta_I$ represents the accumulated offset in the instance dimension.
    }
    \label{tab:rope_convergence}
    \vspace{-10pt}
\end{table}

\subsection{Omni-RoPE: Unified Positional Encoding}
\label{sec:omni_rope}

We introduce \textbf{Omni-RoPE}, a positional encoding scheme tailored for multimodal contexts with complex structural correspondence. Conventional positional encodings cannot reliably distinguish multiple images or preserve spatial alignment across editing operations. Omni-RoPE addresses this limitation by extending Rotary Position Embedding (RoPE)~\citep{su2024roformer} to a unified multimodal setting.

\textbf{Unified formulation.} As shown in Figure~\ref{fig:omni_rope}, each token at coordinates $(h,w)$ in the $k$-th image is assigned a three-dimensional positional identifier:
\begin{equation}
    \operatorname{PosID}_k(h,w) = (\Delta_I^{(k)},\, h,\, w),
\end{equation}
where $\Delta_I^{(k)}$ denotes the \textit{instance identity}, which distinguishes different images or modalities, and $(h,w)$ are the \textit{2D spatial coordinates}. All tokens from the same image share the same $\Delta_I^{(k)}$, while the spatial mapping remains unchanged, i.e., $\mathcal{P}_h^{(k)}(h)=h$ and $\mathcal{P}_w^{(k)}(w)=w$. For text tokens, this formulation naturally reduces to a standard 1D positional index.

This decomposition separates image identity from intra-image spatial layout. Spatial coordinates are computed locally from $(0,0)$ within each image, ensuring that corresponding patches in input and output images receive identical embeddings, thereby preserving spatial alignment and edit consistency. Meanwhile, $\Delta_I^{(k)}$ provides an explicit channel for distinguishing visual instances, which is critical for in-context image generation and multi-image reasoning.

\textbf{Toy experiment verification.}
To evaluate positional correspondence, we design a controlled toy task in which a randomly initialized model is trained to reconstruct the $k$-th image from multiple randomly sampled input images, thereby isolating the effect of positional encoding. We measure efficiency by the number of training steps required to reach a high-fidelity reconstruction target (\textit{loss} $<$ 0.014).

As reported in Table~\ref{tab:rope_convergence}, the RoPE variants used in Lumina-Image-2.0 and Qwen2-VL~\cite{wang2024qwen2} require substantially more training steps to converge, indicating weaker alignment across visual instances. In contrast, \textbf{Omni-RoPE} converges markedly faster and achieves the lowest reconstruction loss, demonstrating stronger spatial correspondence and instance discrimination. Incorporating an \textit{image index embedding}~\cite{chen2025unireal} further improves the final reconstruction fidelity at no additional cost.

\subsection{Foundation Model Training}
We construct the foundation model using a two-stage training pipeline comprising from-scratch pre-training followed by supervised fine-tuning. To accommodate variable context lengths in unified multi-task training, we employ FlashAttention2~\cite{dao2023flashattention} for efficient sequence processing. The model is optimized using the Rectified Flow objective~\cite{liu2022flow,lipman2022flow,albergo2022building}.



\noindent
\textbf{Pre-training.}
This stage focuses on learning general-purpose visual and semantic representations from large-scale datasets. The model is trained through a resolution-based curriculum (256$^2$ $\rightarrow$ 512$^2$ $\rightarrow$ 1024$^2$). For each resolution, we first conduct training on the text-to-image (T2I) task to establish strong text–image alignment. Then, we introduce a curated mixed-task dataset (covering image editing and in-context generation) to diversify the model’s capabilities.

\noindent
\textbf{Supervised Fine-Tuning.}
After pre-training, the model undergoes SFT at 1024$^2$ resolution to refine high-level reasoning and compositional skills. We train on a mixture of curated datasets and distilled data from proprietary models, aiming to enhance instruction following and visual fidelity.

Through two-stage training, the model acquired initial instruction-following skills and versatile generation capability, setting a foundation for subsequent alignment. Detailed configurations for each stage are provided in Appendix 9.7.

\subsection{Instruction Alignment}
We perform online reinforcement learning for multi-task alignment via a progressive curriculum instead of a single joint training stage to ensure stability and avoid task interference. The key challenge is to achieve synergistic gains across tasks without degrading individual performance. 

We define a sequence of training tasks $\mathcal{S} = \langle \mathcal{T}_1, \dots, \mathcal{T}_N \rangle$, where each task $\mathcal{T} = (\tau, \delta, \mathcal{R})$ consists of a \textbf{task type} $\tau \in \{\text{T2I, Edit, IC}\}$, a \textbf{task instance} $\delta$, and a \textbf{reward signal} $\mathcal{R}$. Our goal is to cover all fundamental task types for comprehensive alignment.

For $\tau=\text{Edit}$ and $\tau=\text{IC}$, we adopt general-purpose tasks to enhance instruction-following and compositional abilities. As these tasks lack verifiable rewards, we employ learned reward models: EditScore~\cite{luo2025editscore} for Edit and Qwen2.5-VL-72B~\cite{bai2025qwen25vl} for IC. For $\tau=\text{T2I}$, we select \textbf{GenEval}, which provides verifiable rewards and exhibits strong overlap with Edit and IC. 

We exclude T2I tasks with limited generalization or high reward-hacking risk. In particular, aesthetic rewards such as HPSv3~\cite{ma2025hpsv3} are omitted due to reward hacking, and specialized tasks (e.g., OCR) are excluded as they lack synergy with general instruction-following.

This yields a three-stage curriculum $\langle \mathcal{T}_1, \mathcal{T}_2, \mathcal{T}_3 \rangle$, trained with Flow-GRPO~\cite{liu2025flow}:
\begin{itemize}
    \item $\mathcal{T}_1 = (\text{Edit, general editing, EditScore})$,
    \item $\mathcal{T}_2 = (\text{T2I, GenEval, Verifiable Reward})$,
    \item $\mathcal{T}_3 = (\text{IC, general in-context, Qwen2.5-VL-72B})$.
\end{itemize}

Our RL data includes 50k T2I prompts from Flow-GRPO~\cite{liu2022flow}, 110k editing samples from EditScore~\cite{luo2025editscore}, and 180k in-context data from Echo-4o~\cite{ye2025echo}.

\section{OmniContext Benchmark}


Rigorous evaluation is essential for generalized instruction alignment, particularly for reference-based tasks testing core consistency. However, existing benchmarks fall short, lacking support for multiple input images and diverse tasks. For instance, DreamBench~\cite{ruiz2023dreambooth} only contains 30 objects and 25 prompt templates.  And relying on simplistic metrics like CLIP-I fails on multi-subject evaluation and offers no explainability. To address these critical gaps, we introduce OmniContext, a comprehensive benchmark designed to assess a model's ability to generate content consistent with user-specified context images.

\begin{figure}[t]
  \centering
  \includegraphics[width=1.0\linewidth]{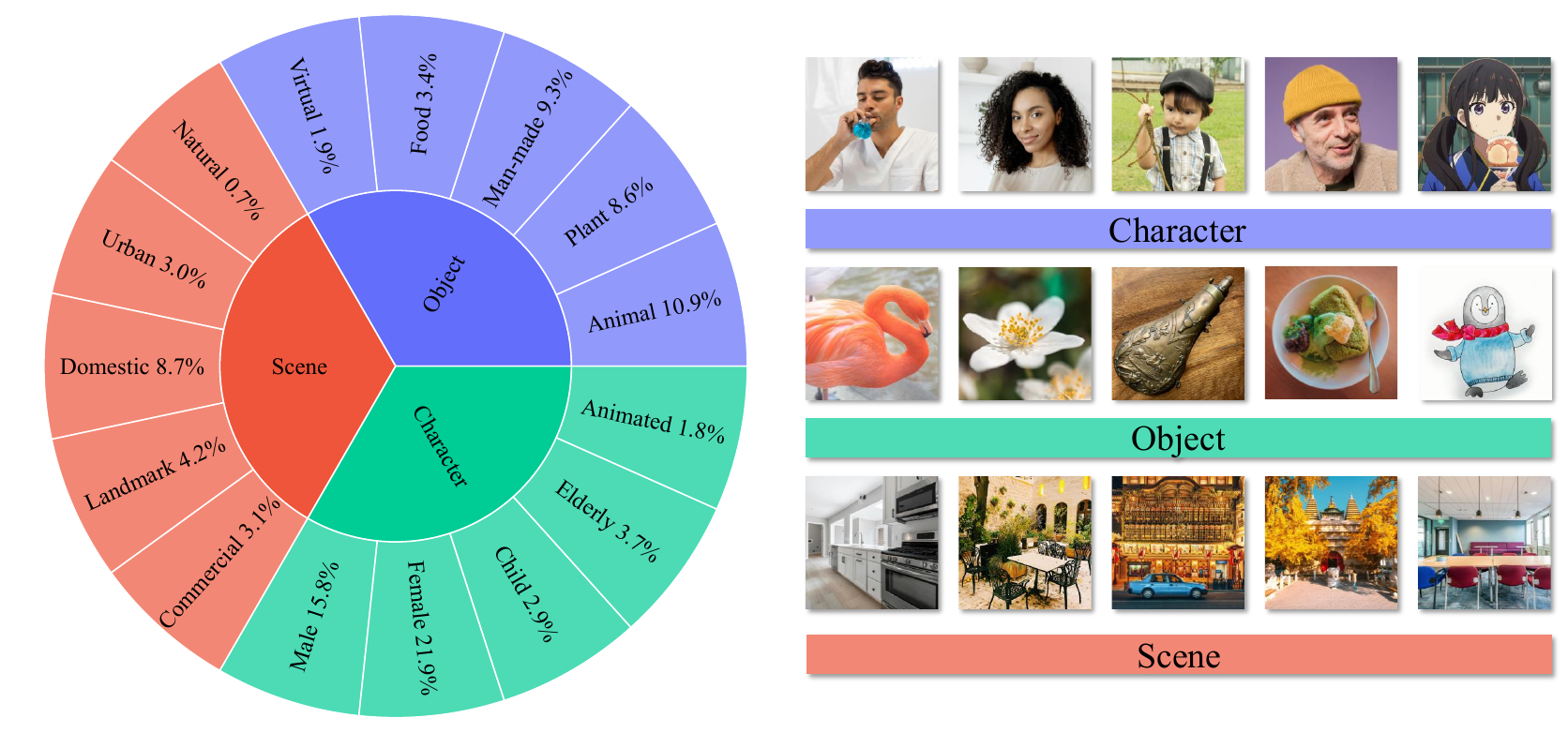}
  \vspace{-10pt}
  \caption{Overview of OmniContext benchmark. \textbf{Left:} Image genres included in OmniContext. \textbf{Right:} Example images for each genre in OmniContext.}
  \label{fig:sunburst} 
  \vspace{-10pt}
\end{figure}

To bridge these gaps, we construct OmniContext using a large-scale, manually collected dataset of high-quality images including personal photos, open-source images, animation stills and AI-generated images. These images are grouped into three distinct categories — Character, Object, and Scene — and exhibit diverse coverage across various domains, as illustrated in Figure~\ref{fig:sunburst}. We define three task categories(SINGLE, MULTIPLE, and SCENE), each with 50 examples per subtask. SINGLE uses one context image, MULTIPLE combines multiple subjects, and SCENE conditions on environmental context.

Image–prompt pairs are constructed through a hybrid process combining MLLMs and manual annotation. MLLMs first filter low-quality samples, after which experts select images based on clarity, aesthetics, and diversity. Prompts are generated with GPT-4o and refined for semantic and syntactic variety.

We use GPT-4.1 to assess outputs on three metrics: Prompt Following (PF), Subject Consistency (SC), and an Overall Score (geometric mean of PF and SC). Following VIEScore~\cite{ku2023viescore}, GPT-4.1 provides both scores (0–10) and rationales to justify its evaluations. We believe the OmniContext will serve as a valuable resource for driving future research in controllable, reference-based image generation.

\section{Experiments}

\begin{table*}[t]
    \centering
    \resizebox{0.99\textwidth}{!}{
        \begin{tabular}{lcccc|cc|cc|ccc}
        \toprule
        \textbf{Model} & \# Params & \multicolumn{3}{c|}{\textbf{Understanding}} & \multicolumn{2}{c|}{\textbf{Image Generation}} & \multicolumn{2}{c|}{\textbf{Image Editing}} & \multicolumn{3}{c}{\textbf{In-context Generation}} \\
        
        & & MMB$\uparrow$ & MMMU$\uparrow$ & MM-Vet$\uparrow$ & GenEval$\uparrow$ & OneIG-Bench-EN$\uparrow$ & ImgEdit-Bench$\uparrow$ & GEdit-Bench-EN$\uparrow$ & Single$\uparrow$ & Multiple$\uparrow$ & Scene$\uparrow$ \\
        \midrule
        LLaVA-1.5~\cite{llava} & - & 36.4 & 67.8 & 36.3 & - & - & - & - & - & - & -\\
        LLaVA-NeXT~\cite{liu2024llavanext} & - & 79.3 & 51.1 & 57.4 & - & - & - & - & - & - & - \\
        \midrule
        SDXL~\cite{podell2023sdxl} & - & - & - & - & 0.55 & 0.32 & - & - & - & - & - \\
        SD3-medium~\cite{sd3-medium} & - & - & - & - & 0.62 & - & - & - & - & - & - \\
        FLUX.1-dev~\cite{FLUX} & - & - & - & - & 0.66 & 0.43 & - & - & - & - & - \\
        Qwen-Image~\cite{wu2025qwenimagetechnicalreport} & - & - & - & - & \underline{0.91} & \underline{0.54} & - & - & - & - & - \\
        \midrule
        Instruct-P2P~\cite{brooks2023instructpix2pix} & - & - & - & - & - & - & 1.88 & 3.68 & - & - & - \\
        MagicBrush~\cite{zhang2024magicbrush} & - & - & - & - & - & - & 1.90 & 1.86 & - & - & - \\
        AnyEdit~\cite{yu2025anyedit} & - & - & - & - & - & - & 2.45 & 3.21 & - & - & - \\
        Step1X-Edit~\cite{liu2025step1x} & - & - & - & - & - & - & 3.06 & 6.70 & - & - & - \\
        IC-Edit~\cite{zhang2025context} & - & - & - & - & - & - & 3.05 & 4.84 & - & - & - \\
        \midrule
        UNO~\cite{uno} & - & - & - & - & - &- & - & - & 6.72 & 4.48  & 3.59 \\
        InfiniteYou~\cite{jiang2025infiniteyou} & - & - & - & - & - &- & - & - & 6.05 & - & - \\
        DreamO~\cite{mou2025dreamo} & - & - & - & - & - &- & - & - & 7.65 & 7.05 & 4.52 \\
        UMO~\cite{cheng2025umo} & - & - & - & - & - &- & - & - & 7.78 & 7.14 & 6.78 \\
        \midrule
        Show-o~\cite{xie2024show} & - & - & 27.4 & - & 0.68 & - & - & - & - & - & - \\
        Janus-Pro~\cite{chen2025janus} & - & 75.5 & 36.3 & 39.8 & 0.80 & 0.26 & - & - & - & - & - \\
        Emu3~\cite{wang2024emu3} & - & 58.5 & 31.6 & 37.2 & 0.54 / $0.66^{\dagger}$ & - & - & - & - & - & - \\
        MetaQuery-XL~\cite{pan2025metaquery} & 7B + 1.6B$^{*}$ & \underline{83.5} & \textbf{58.6} & 66.6 & $0.80^{\dagger}$ & -  & - & - & - & - & - \\
        BLIP3-o 8B~\cite{chen2025blip3} & 7B + 1.4B$^{*}$ & \underline{83.5} & \textbf{58.6} & 66.6 & $0.84^{\dagger}$ & 0.31 &  - & - & - & - & - \\
        BAGEL~\cite{bagel} & 7B + 7B$^{*}$ & \textbf{85.0} & 55.3 & 67.2 & 0.82 / 0.88$^{\dagger}$ & 0.36 & 3.20 & 6.52 & 6.25 & 6.02 & 5.08 \\
        UniWorld-V1~\cite{lin2025uniworld} & 7B + 12B$^{*}$ & \underline{83.5} & \textbf{58.6} & 67.1 & $0.84^{\dagger}$ & - & 3.26 & 4.85 & - & - & - \\
        \midrule
        Qwen-Image-Edit-2509~\cite{wu2025qwenimagetechnicalreport} & 7B + 20B & - & - & - & - & - & \textbf{4.41} & \textbf{7.54} & \underline{8.74} & \textbf{8.13} & 6.55 \\
        Gemini 2.5 Flash Image~\cite{google2025gemini25flash} & - & - & - & - & - & \textbf{0.55} & \underline{4.28} & 7.10 & \textbf{8.77} & \underline{8.06} & \underline{7.01} \\
        \midrule
        OmniGen~\cite{xiao2025omnigen} & 3.8B & - & - & - & 0.68 & - & 2.96 & 5.06 & 6.46 & 5.26 & 4.34 \\
         \rowcolor{myblue}\textbf{OmniGen2} & 3B + 4B$^{*}$ & 79.1 & 53.1 & 61.8 & \textbf{0.95}  & 0.47 & 3.69 & \underline{7.21} & 8.41 & 7.73 & \textbf{7.86} \\
        \bottomrule
        \end{tabular}
    }
    \vspace{5pt}
     \caption{Comparison of different models across Understanding, Generation, Editing, and In-context Generation tasks. *: The first term represents the number of parameters for text generation, while the second term corresponds to the number of parameters allocated for image generation. $\dagger$ refers to the methods using LLM rewriter.}
    \label{tab:main_comparison}
    \vspace{-10pt}
\end{table*}

In this section, we conduct a comprehensive evaluation of OmniGen2 to demonstrate its unified capabilities across a wide spectrum of generation tasks. The overall comparison results are presented in Table~\ref{tab:main_comparison}.


\subsection{Visual Understanding}
\label{subsec:understanding}

OmniGen2 leverages Qwen2.5-VL-3B-Instruct~\cite{bai2025qwen25vl} for visual understanding. As shown in Table~\ref{tab:main_comparison}, our model demonstrates robust multimodal comprehension, achieving solid scores of 79.1 on MMBench~\cite{liu2024mmbench}, 53.1 on MMMU~\cite{yue2024mmmu}, and 61.8 on MM-Vet~\cite{yu2023mm}. These results confirm that OmniGen2 possesses a solid foundation for interpreting complex visual and textual instructions, which is essential for high-quality, instruction-aligned generation.

\subsection{Text-to-Image Generation}
\label{subsec:t2i}

We assess OmniGen2's T2I generation capabilities on two standard benchmarks: GenEval~\cite{ghosh2023geneval}, which evaluates compositional generation, and OneIG-Bench~\cite{chang2025oneig} which evaluate models across multiple dimensions, including prompt-image alignment, text rendering precision, reasoning-generated content, stylization, and diversity. 

As shown in Table \ref{tab:main_comparison}, OmniGen2 delivers strong image generation performance on complex, compositional prompts, achieving an overall score of \textbf{0.95} on GenEval. This surpasses other powerful unified models such as UniWorld-V1 (0.84) and BAGEL (0.88), as well as Qwen-Image, a model specialized for T2I generation, highlighting the effectiveness of the RL alignment strategy in OmniGen2. On the more comprehensive OneIG-Bench, OmniGen2 continues to demonstrate competitive realism, achieving an overall score of 0.47, outperforming most existing models and trailing only behind large-scale models such as Gemini 2.5 Flash Image and Qwen-Image. More details are provided in Appendix 8.1 and 9.10.




\subsection{Image Editing}
\label{subsec:editing}
\begin{table}[t]
\centering
\resizebox{0.99\linewidth}{!}{
    \begin{tabular}{lcccccc}
        \toprule
        Method & \multicolumn{3}{c}{Emu-Edit} & \multicolumn{3}{c}{GEdit-Bench-EN} \\
        & \multicolumn{1}{c}{CLIP-I$\uparrow$} & \multicolumn{1}{c}{CLIP-Out$\uparrow$} & \multicolumn{1}{c}{DINO$\uparrow$} & \multicolumn{1}{c}{\space SC$\uparrow$} & \multicolumn{1}{c}{\space PQ$\uparrow$} & \multicolumn{1}{c}{\space O $\uparrow$} \\
        \midrule
        Gemini-2.0-Flash-Image~\cite{gemini-2.0-flash} & - & - & - & 6.73 & 6.61 & 6.32 \\
         Gemini-2.5-Flash-Image~\cite{google2025gemini25flash} & - & - & - & 7.41 & 7.96 & 7.10 \\
        GPT-4o~\cite{gpt4o} & - & - & - & 7.85 & 7.62 & 7.53 \\
        \midrule
        Instruct-Pix2Pix~\cite{brooks2023instructpix2pix} & 0.856 & 0.292 & 0.773 & 3.58 & 5.49 & 3.68 \\
        MagicBrush~\cite{zhang2024magicbrush} & 0.877 & 0.298 & 0.807 & 4.68 & 5.66 & 4.52 \\
        AnyEdit~\cite{yu2025anyedit} & - & - & - & 3.18 & 5.82 & 3.21 \\
        OmniGen~\cite{xiao2025omnigen} & - & - & - & 5.96 & 5.89 & 5.06 \\
        ICEdit~\cite{zhang2025context} & \textbf{0.907} & 0.305 & \underline{0.866} & 5.11 & 6.85 & 4.84   \\
        Step1X-Edit~\cite{liu2025step1x} & 0.860 & 0.304 & 0.782 & 7.09 & 6.76 & 6.70 \\
        BAGEL~\cite{bagel} & 0.839 & \underline{0.307} & 0.753 & 7.36 & 6.83 & 6.52 \\
        UniWorld-V1~\cite{lin2025uniworld} & - & - & - & 4.93 & 7.43 & 4.85 \\
        Qwen-Image-Edit-2509~\cite{wu2025qwenimagetechnicalreport} & - & - & - & \textbf{8.15} & \underline{7.86} & \textbf{7.54} \\
        \rowcolor{myblue}\textbf{OmniGen2} & \underline{0.896} & \textbf{0.311} & \textbf{0.876} & \underline{7.58} & \textbf{7.94} & \underline{7.21} \\
        \bottomrule
    \end{tabular}
}
\caption{Quantitative comparison on Emu-Edit~\cite{sheynin2024emuedit} and GEdit-Bench-EN~\cite{liu2025step1x}. For Emu-Edit, CLIP-I/DINO measure consistency with the source image, while CLIP-Out measures alignment with the caption of target image, CLIP-B/32~\cite{radford2021learning} and DINO-S/16~\cite{caron2021emerging} are leveraged for feature calculation. For GEdit-Bench, SC (Semantic Consistency) evaluates instruction following, and PQ (Perceptual Quality) assesses image naturalness and artifacts. Higher scores are better for all metrics.}
\vspace{-5pt}
\label{tab:gedit_emu}
\end{table}

Image editing is a cornerstone of OmniGen2's capabilities. We rigorously evaluate its performance across three diverse benchmarks: Emu-Edit~\cite{sheynin2024emu}, GEdit-Bench-EN~\cite{liu2025step1x} and ImgEdit-Bench~\cite{ye2025imgedit}. The results collectively demonstrate that OmniGen2 achieve a strong performance in instruction-based image editing. 

As shown in Table \ref{tab:gedit_emu}, OmniGen2 exhibits an exceptional balance between edit accuracy and image preservation. On Emu-Edit, our model achieves the highest CLIP-Out score~(0.311), indicating it most effectively applies the requested edits among all compared models. Concurrently, it secures the second-best scores for CLIP-I~(0.896) and best scores for DINO~(0.876), which measure the preservation of unedited regions. This combination highlights OmniGen2's proficiency in making precise, localized changes without disturbing the rest of the image.
This strong instruction-following capability is further confirmed on GEdit-Bench, where OmniGen2 achieves the second-highest Semantic Consistency (SC) score of 7.58 and the highest Perceptual Quality (PQ) score of 7.94. This leads to a strong overall score of 7.21, placing it among the top-tier models. This score \textbf{outperforms} Gemini-2.5-Flash-Image and is \textbf{second only} to Qwen-Image-Edit among open-source models. 
As detailed in Table~\ref{tab:main_comparison}, OmniGen2 demonstrates compelling performance on the comprehensive ImgEdit-Bench, notably surpassing some strong open-source models like BAGEL. More details are provided in Appendix 8.2.

\begin{table*}[t]
    \centering
    \resizebox{0.99\linewidth}{!}{
    \begin{tabular}{l|cc|ccc|ccc|c}
        \toprule
        \multirow{2}{*}{Method} & \multicolumn{2}{c|}{SINGLE} & \multicolumn{3}{c|}{MULTIPLE} & \multicolumn{3}{c|}{SCENE} & \multirow{2}{*}{Average$\uparrow$}\\ 
        \cmidrule(lr){2-9}
        & Character & Object & Character & Object & Char. + Obj. & Character & Object & Char. + Obj. & \\
        \midrule
        Flux.1 Kontext max~\cite{fluxkontext} & 8.48 & 8.68 & - & - & - & - & - & - & - \\
        Gemini-2.0-Flash-Image~\cite{gemini-2.0-flash} & 5.06 & 5.17 & 2.91 & 2.16 & 3.80 & 3.02 & 3.89 & 2.92 & 3.62\\
        Gemini-2.5-Flash-Image~\cite{google2025gemini25flash} & 8.62 & 8.91 & 7.88 & 8.92 & 7.39 & 7.29 & 7.05 & 6.68 & 7.84\\
        GPT-4o~\cite{gpt4o} & \textbf{8.90} & \textbf{9.01} & \textbf{9.07} & \textbf{8.95} & \textbf{8.54} & \textbf{8.90} & \textbf{8.44} & \textbf{8.60} & \textbf{8.80} \\
        \midrule
        InfiniteYou~\cite{jiang2025infiniteyou} & 6.05 & - & - & - & - & - & - & - & - \\
        UNO~\cite{uno} & 6.60 & 6.83 & 2.54 & 6.51 & 4.39 & 2.06 & 4.33 & 4.37 & 4.71\\
        BAGEL~\cite{bagel} & 5.48 & 7.03 & 5.17 & 6.64 & 6.24 & 4.07 & 5.71 & 5.47 & 5.73\\
         Qwen-Image-Edit-2509~\cite{wu2025qwenimagetechnicalreport} & \textbf{8.35} & \textbf{9.13} & \textbf{7.65} & \textbf{8.85} & 7.90 & 5.16 & 7.75 & 6.73 & 7.69 \\
        OmniGen~\cite{xiao2025omnigen} & 7.21 & 5.71 & 5.65 & 5.44 & 4.68 & 3.59 & 4.32 & 5.12 & 4.34\\
         \rowcolor{myblue}\textbf{OmniGen2} & 8.19 & 8.63 & 7.45 & 7.91 & \textbf{7.93} & \textbf{7.75} & \textbf{7.91} & \textbf{7.93} & \textbf{7.95} \\
        \bottomrule
    \end{tabular}
}
\caption{Overall comparison of existing models on our proposed OmniContext benchmark. "Char. + Obj." indicates Character + Object.}
\label{tab:omni_context}
\vspace{-12pt}
\end{table*}

\subsection{In-context Generation}
\label{subsec:in_context}

A distinguishing feature of OmniGen2 is its capacity to perform in-context generation.
We introduce the \textbf{OmniContext} benchmark to provide a comprehensive evaluation of the performance of the existing model in this domain. OmniContext comprises eight subtasks, with overall scores for each subtask presented in Table~\ref{tab:omni_context}. As the inaugural model evaluated on this benchmark, OmniGen2 establishes a strong baseline, achieving an overall score of 7.95, which \textbf{surpass} the powerful open-sourced model Qwen-Image-Edit-2509. These results show OmniGen2's proficiency in disentangling the subject's identity from its original background and re-rendering it accurately according to new textual instructions. OmniGen2 exhibits significant improvements over competing models in all types of tasks, demonstrating superior prompt-following ability and subject consistency. Among closed-source models, GPT-4o~\cite{gpt4o} achieves the highest scores in the Overall metrics. More details are provided in Appendix 8.3, 9.8. 

\subsection{Ablation Study}
Our ablation study, detailed in Table~\ref{tab:rl_ablation}, validates our principled curriculum by demonstrating the critical importance of two key factors: the \textbf{selection} of tasks and reward signals, and the \textbf{scheduling} of the training sequence. For task selection, we highlight four crucial findings: (1) Tasks with limited skill overlap can cause negative interference, as shown by \texttt{OCR only} training which degrades the GEdit Overall score from 6.28 to 6.13. (2) Conversely, well-chosen tasks with skill overlap such as instruction following exhibit strong synergy; the \texttt{Edit \& GenEval} strategy surpasses both single-task baselines on their respective metrics (GenEval: 0.95 vs. 0.94; GEdit Overall: 7.19 vs. 7.01). (3) Reward signals about human preference pose significant risks, with \texttt{Edit \& HPSv3} confirming reward hacking by inflating the PQ score to 8.22 at the severe cost of collapsing SC and IC scores. (4) accuracy reward signal is vital. As shown by \texttt{Edit only}, whose IC score is higher than \texttt{IC only} (7.71 vs. 7.38) because of excel performance of EditScore~\cite{luo2025editscore} to enhance instruction following. Beyond selection, the training sequence is equally vital. This is confirmed by comparing our final curriculum (\texttt{Edit \& Geneval \& IC}) against an alternative ordering (\texttt{Edit \& IC \& Geneval}), which results in a marked performance drop (GEdit Overall: 7.21 vs. 7.06). We also observe that prioritizing editing tasks leads to consistently better performance than T2I-first. We hypothesize this is because editing tasks with richer supervision build a robust foundation for subsequent learning. And Additional results on out-of-distribution (OOD) benchmarks are provided in Appendix 9.10, showing consistent improvements under our RL curriculum. These findings collectively prove that both careful selection and scheduling are essential to our principled alignment strategy.
\begin{table}[ht!]
    \centering
    \small 
    \setlength{\tabcolsep}{3.5pt} 
    \resizebox{0.99\linewidth}{!}{
    \begin{tabular}{l c c ccc} 
        \toprule
        \multirow{2}{*}{Strategy} & \multirow{2}{*}{GenEval $\uparrow$} & \multirow{2}{*}{OmniContext $\uparrow$} & \multicolumn{3}{c}{GEdit $\uparrow$} \\
        \cmidrule(lr){4-6}
        & & & SC & PQ & Overall \\
        \midrule
        Base (w/o RL) & 0.78 & 7.18 & 6.72 & 7.20 & 6.28 \\
        \midrule
        \multicolumn{6}{l}{\textit{Single-Task}} \\ 
        Edit only & 0.79 & 7.71 & 7.30 & 7.95 & 7.01 \\
        GenEval only & 0.94 & 7.24 & 6.78 & 7.20 & 6.30 \\
        OCR only & 0.78 & 7.33 & 6.65 & 7.15 & 6.13 \\
        IC only & 0.78 & 7.38 & 6.97 & 6.98 & 6.39 \\
        \midrule
        \multicolumn{6}{l}{\textit{Multi-Tasks}} \\ 
        Edit \& GenEval & \textbf{0.95} & 7.68 & 7.52 & 7.95 & 7.19 \\
        Edit \& OCR & 0.81 & 7.70 & 7.28 & 7.96 & 7.06 \\
        Edit \& HPSv3 & 0.77 & 6.82 & 6.87  & \textbf{8.22} & 6.88 \\
        Edit \& IC \& GenEval & 0.93 & 7.65 & 7.33  & 7.92 & 7.06 \\
        Geneval \& Edit \& IC & 0.94 & 7.80 & 7.49  & 7.97 & \textbf{7.21} \\
        \rowcolor{myblue}\textbf{Edit \& GenEval \& IC (Ours)}  & \textbf{0.95} & \textbf{7.95} & \textbf{7.58}  & 7.94 & \textbf{7.21} \\
        \bottomrule
    \end{tabular}
    }
    \caption{Ablation study of multi-task reinforcement learning strategies. T2I, Edit, and IC tasks are trained for 1500, 700, and 200 steps, respectively. }
    \label{tab:rl_ablation}
\end{table}
\section{Related Works}

\subsection{Multimodal Generation}
Recent advances in multimodal generation have produced models capable of both understanding and generating content across text, images, and video. Diffusion-based models, including the Stable Diffusion series~\cite{LDM,podell2023sdxl,sd3}, DALL·E~\cite{dalle2}, and Imagen~\cite{imagen3}, have achieved high-fidelity image synthesis, while methods like ControlNet~\cite{controlnet} and T2I-Adapter~\cite{t2iadapter} improve controllability, and StyleShot, InstructPix2Pix, and EMU-Edit~\cite{gao2024styleshot,brooks2023instructpix2pix,sheynin2024emuedit} support fine-grained, instruction-guided editing. Unified image generation models such as OmniGen~\cite{xiao2025omnigen}, UniReal~\cite{chen2025unireal}, and related works~\cite{xiao2025omnigen,mao2025ace++,chen2025unireal,tian2025mige} extend this further, integrating multiple tasks into a single model.
Building on this foundation, autoregressive multimodal models provide an alternative paradigm for unified generation~\cite{team2024chameleon,sun2024generative,wang2024emu3}.
There are also hybrid approaches such as Show-o and Transfusion~\cite{xie2024show,zhou2024transfusion,gupta2022metamorph,dong2023dreamllm,shi2024llamafusion,pan2025metaquery} combining autoregressive text generation with diffusion-based image modeling. Several works focus on adapting large language models for multimodal generation~\cite{chen2025blip3,liao2025mogao,bagel,wang2025ovisu1technicalreport,wu2025qwen}: These works are trained with vast amount of data, obtaining powerful image understanding and image generation capabilities.

\subsection{Reinforcement Learning in Diffusion Model}
Reinforcement learning has increasingly been adopted to improve alignment in diffusion and flow-based models.
For text-to-image (T2I) generation, early works such as DDPO and DPOK~\cite{black2024trainingdiffusionmodelsreinforcement,fan2023dpok} formulated diffusion sampling as a sequential decision process and optimized via KL-regularized policy updates. 
Follow-up approaches including ReFL, AlignProp~\cite{chen2024enhancing,prabhudesai2023aligning} refined this paradigm with more stable reward optimization, improved credit assignment across denoising steps, and scalable preference-learning from human or synthetic feedback. More recently GRPO~\cite{shao2024deepseekmath} has become prominent due to its training stability and efficiency.
GRPO-based alignment method like DanceGRPO~\cite{xue2025dancegrpo}, Flow-GRPO~\cite{liu2025flow}, and Mix-GRPO~\cite{li2025mixgrpounlockingflowbasedgrpo} further push the boundaries of alignment technology, outperforming traditional methods in both accuracy and scalability~\cite{he2025tempflow,wang2025pref,li2025branchgrpo}. 
For image editing or in-context generation, RL has also been used to enforce text alignment and editing faithfulness to ensure consistency between input and output~\cite{huang2025competition,miao2024subject,luo2025editscore,cheng2025umo}. 
Despite the rapid progress, most existing approaches optimize RL for a single task or a narrow alignment objective. In contrast, our work introduces a multi-task RL pipeline that jointly aligns the model's behavior across all three critical scenarios, achieving comprehensive, all-around alignment.
\section{Conclusion}
In this work, we present OmniGen2, a generative model that is systematically instruction aligned. OmniGen2 explores two directions to enhance alignment performance: constructing a robust and flexible base model, and developing a multi-task RL alignment scheme. OmniGen2 utilizes a simple, efficient and flexible architecture to support diverse multimodal generation tasks. Our experiments across standard benchmarks and our propose novel OmniContext benchmark demonstrate OmniGen2's semantic consistency, versatile capabilities, and superior generation quality. Instruction alignment has consistently and significantly enhanced the base model across various tasks. These results suggest that instruction alignment may represent a crucial step toward realizing general multimodal systems.

{
    \small
    \bibliographystyle{ieeenat_fullname}
    \bibliography{main}

@String(ICCV= {Int. Conf. Comput. Vis.})

@String(ICLR = {Int. Conf. Learn. Represent.})

@String(ICCV  = {ICCV})

@String(ICLR  = {ICLR})

@InProceedings{controlnet,
    author    = {Zhang, Lvmin and Rao, Anyi and Agrawala, Maneesh},
    title     = {Adding Conditional Control to Text-to-Image Diffusion Models},
    booktitle = {Proceedings of the IEEE/CVF International Conference on Computer Vision (ICCV)},
    month     = {October},
    year      = {2023},
    pages     = {3836-3847}
}

@article{llava,
  title={Visual instruction tuning},
  author={Liu, Haotian and Li, Chunyuan and Wu, Qingyang and Lee, Yong Jae},
  journal={Advances in neural information processing systems},
  volume={36},
  year={2024}
}

@inproceedings{LDM,
  title={High-resolution image synthesis with latent diffusion models},
  author={Rombach, Robin and Blattmann, Andreas and Lorenz, Dominik and Esser, Patrick and Ommer, Bj{\"o}rn},
  booktitle={Proceedings of the IEEE/CVF conference on computer vision and pattern recognition},
  pages={10684--10695},
  year={2022}
}

@article{podell2023sdxl,
  title={Sdxl: Improving latent diffusion models for high-resolution image synthesis},
  author={Podell, Dustin and English, Zion and Lacey, Kyle and Blattmann, Andreas and Dockhorn, Tim and M{\"u}ller, Jonas and Penna, Joe and Rombach, Robin},
  journal={arXiv preprint arXiv:2307.01952},
  year={2023}
}

@inproceedings{sd3,
  title={Scaling rectified flow transformers for high-resolution image synthesis},
  author={Esser, Patrick and Kulal, Sumith and Blattmann, Andreas and Entezari, Rahim and M{\"u}ller, Jonas and Saini, Harry and Levi, Yam and Lorenz, Dominik and Sauer, Axel and Boesel, Frederic and others},
  booktitle={Forty-first International Conference on Machine Learning},
  year={2024}
}

@article{dalle2,
  title={Hierarchical text-conditional image generation with clip latents},
  author={Ramesh, Aditya and Dhariwal, Prafulla and Nichol, Alex and Chu, Casey and Chen, Mark},
  journal={arXiv preprint arXiv:2204.06125},
  volume={1},
  number={2},
  pages={3},
  year={2022}
}

@misc{imagen3,
      title={Imagen 3}, 
      author={Imagen-Team-Google},
      year={2024},
      eprint={2408.07009},
      archivePrefix={arXiv},
      primaryClass={cs.CV},
      url={https://arxiv.org/abs/2408.07009}, 
}

@misc{t2iadapter,
      title={T2I-Adapter: Learning Adapters to Dig out More Controllable Ability for Text-to-Image Diffusion Models}, 
      author={Chong Mou and Xintao Wang and Liangbin Xie and Yanze Wu and Jian Zhang and Zhongang Qi and Ying Shan and Xiaohu Qie},
      year={2023},
      eprint={2302.08453},
      archivePrefix={arXiv},
      primaryClass={cs.CV},
      url={https://arxiv.org/abs/2302.08453}, 
}

@inproceedings{brooks2023instructpix2pix,
  title={Instructpix2pix: Learning to follow image editing instructions},
  author={Brooks, Tim and Holynski, Aleksander and Efros, Alexei A},
  booktitle={Proceedings of the IEEE/CVF Conference on Computer Vision and Pattern Recognition},
  pages={18392--18402},
  year={2023}
}

@article{ye2023ipadapter,
  title={Ip-adapter: Text compatible image prompt adapter for text-to-image diffusion models},
  author={Ye, Hu and Zhang, Jun and Liu, Sibo and Han, Xiao and Yang, Wei},
  journal={arXiv preprint arXiv:2308.06721},
  year={2023}
}

@inproceedings{ruiz2023dreambooth,
  title={Dreambooth: Fine tuning text-to-image diffusion models for subject-driven generation},
  author={Ruiz, Nataniel and Li, Yuanzhen and Jampani, Varun and Pritch, Yael and Rubinstein, Michael and Aberman, Kfir},
  booktitle={Proceedings of the IEEE/CVF conference on computer vision and pattern recognition},
  pages={22500--22510},
  year={2023}
}

@article{gao2024styleshot,
  title={StyleShot: A Snapshot on Any Style},
  author={Gao, Junyao and Liu, Yanchen and Sun, Yanan and Tang, Yinhao and Zeng, Yanhong and Chen, Kai and Zhao, Cairong},
  journal={arXiv preprint arXiv:2407.01414},
  year={2024}
}

@article{liu2022flow,
  title={Flow straight and fast: Learning to generate and transfer data with rectified flow},
  author={Liu, Xingchao and Gong, Chengyue and Liu, Qiang},
  journal={arXiv preprint arXiv:2209.03003},
  year={2022}
}

@article{chen2023pixart,
  title={Pixart-alpha: Fast training of diffusion transformer for photorealistic text-to-image synthesis},
  author={Chen, Junsong and Yu, Jincheng and Ge, Chongjian and Yao, Lewei and Xie, Enze and Wu, Yue and Wang, Zhongdao and Kwok, James and Luo, Ping and Lu, Huchuan and others},
  journal={arXiv preprint arXiv:2310.00426},
  year={2023}
}

@article{li2024recaption,
      title={What If We Recaption Billions of Web Images with LLaMA-3?}, 
      author={Xianhang Li and Haoqin Tu and Mude Hui and Zeyu Wang and Bingchen Zhao and Junfei Xiao and Sucheng Ren and Jieru Mei and Qing Liu and Huangjie Zheng and Yuyin Zhou and Cihang Xie},
      journal={arXiv preprint arXiv:2406.08478},
      year={2024}
}

@article{chen2023sharegpt4v,
  title={Sharegpt4v: Improving large multi-modal models with better captions},
  author={Chen, Lin and Li, Jisong and Dong, Xiaoyi and Zhang, Pan and He, Conghui and Wang, Jiaqi and Zhao, Feng and Lin, Dahua},
  journal={arXiv preprint arXiv:2311.12793},
  year={2023}
}

@article{schuhmann2022laion,
  title={Laion-5b: An open large-scale dataset for training next generation image-text models},
  author={Schuhmann, Christoph and Beaumont, Romain and Vencu, Richard and Gordon, Cade and Wightman, Ross and Cherti, Mehdi and Coombes, Theo and Katta, Aarush and Mullis, Clayton and Wortsman, Mitchell and others},
  journal={Advances in Neural Information Processing Systems},
  volume={35},
  pages={25278--25294},
  year={2022}
}

@misc{chen2024allava,
      title={ALLaVA: Harnessing GPT4V-synthesized Data for A Lite Vision-Language Model}, 
      author={Guiming Hardy Chen and Shunian Chen and Ruifei Zhang and Junying Chen and Xiangbo Wu and Zhiyi Zhang and Zhihong Chen and Jianquan Li and Xiang Wan and Benyou Wang},
      year={2024},
      eprint={2402.11684},
      archivePrefix={arXiv},
      primaryClass={cs.CL}
}

@article{onoe2024docci,
  title={DOCCI: Descriptions of Connected and Contrasting Images},
  author={Onoe, Yasumasa and Rane, Sunayana and Berger, Zachary and Bitton, Yonatan and Cho, Jaemin and Garg, Roopal and Ku, Alexander and Parekh, Zarana and Pont-Tuset, Jordi and Tanzer, Garrett and others},
  journal={arXiv preprint arXiv:2404.19753},
  year={2024}
}

@article{li2024DenseFusion,
      title={DenseFusion-1M: Merging Vision Experts for Comprehensive Multimodal Perception}, 
      author={Xiaotong Li and Fan Zhang and Haiwen Diao and Yueze Wang and Xinlong Wang and Ling-Yu Duan},
      year={2024},
      journal={2407.08303}
}

@article{sun2024journeydb,
  title={Journeydb: A benchmark for generative image understanding},
  author={Sun, Keqiang and Pan, Junting and Ge, Yuying and Li, Hao and Duan, Haodong and Wu, Xiaoshi and Zhang, Renrui and Zhou, Aojun and Qin, Zipeng and Wang, Yi and others},
  journal={Advances in Neural Information Processing Systems},
  volume={36},
  year={2024}
}

@article{ge2024seededit,
  title={SEED-Data-Edit Technical Report: A Hybrid Dataset for Instructional Image Editing},
  author={Ge, Yuying and Zhao, Sijie and Li, Chen and Ge, Yixiao and Shan, Ying},
  journal={arXiv preprint arXiv:2405.04007},
  year={2024}
}

@article{zhang2024magicbrush,
  title={Magicbrush: A manually annotated dataset for instruction-guided image editing},
  author={Zhang, Kai and Mo, Lingbo and Chen, Wenhu and Sun, Huan and Su, Yu},
  journal={Advances in Neural Information Processing Systems},
  volume={36},
  year={2024}
}

@inproceedings{rombach2022high,
  title={High-resolution image synthesis with latent diffusion models},
  author={Rombach, Robin and Blattmann, Andreas and Lorenz, Dominik and Esser, Patrick and Ommer, Bj{\"o}rn},
  booktitle={Proceedings of the IEEE/CVF conference on computer vision and pattern recognition},
  pages={10684--10695},
  year={2022}
}

@article{ghosh2024geneval,
  title={Geneval: An object-focused framework for evaluating text-to-image alignment},
  author={Ghosh, Dhruba and Hajishirzi, Hannaneh and Schmidt, Ludwig},
  journal={Advances in Neural Information Processing Systems},
  volume={36},
  year={2024}
}

@inproceedings{sheynin2024emuedit,
  title={Emu edit: Precise image editing via recognition and generation tasks},
  author={Sheynin, Shelly and Polyak, Adam and Singer, Uriel and Kirstain, Yuval and Zohar, Amit and Ashual, Oron and Parikh, Devi and Taigman, Yaniv},
  booktitle={Proceedings of the IEEE/CVF Conference on Computer Vision and Pattern Recognition},
  pages={8871--8879},
  year={2024}
}

@article{groundingdino,
  title={Grounding dino: Marrying dino with grounded pre-training for open-set object detection},
  author={Liu, Shilong and Zeng, Zhaoyang and Ren, Tianhe and Li, Feng and Zhang, Hao and Yang, Jie and Li, Chunyuan and Yang, Jianwei and Su, Hang and Zhu, Jun and others},
  journal={arXiv preprint arXiv:2303.05499},
  year={2023}
}

@article{lightman2023let,
  title={Let's verify step by step},
  author={Lightman, Hunter and Kosaraju, Vineet and Burda, Yura and Edwards, Harri and Baker, Bowen and Lee, Teddy and Leike, Jan and Schulman, John and Sutskever, Ilya and Cobbe, Karl},
  journal={arXiv preprint arXiv:2305.20050},
  year={2023}
}

@article{jaech2024openai,
  title={Openai o1 system card},
  author={Jaech, Aaron and Kalai, Adam and Lerer, Adam and Richardson, Adam and El-Kishky, Ahmed and Low, Aiden and Helyar, Alec and Madry, Aleksander and Beutel, Alex and Carney, Alex and others},
  journal={arXiv preprint arXiv:2412.16720},
  year={2024}
}

@article{guo2025deepseek,
  title={DeepSeek-R1 incentivizes reasoning in LLMs through reinforcement learning},
  author={Guo, Daya and Yang, Dejian and Zhang, Haowei and Song, Junxiao and Wang, Peiyi and Zhu, Qihao and Xu, Runxin and Zhang, Ruoyu and Ma, Shirong and Bi, Xiao and others},
  journal={Nature},
  volume={645},
  number={8081},
  pages={633--638},
  year={2025},
  publisher={Nature Publishing Group UK London}
}

@article{hurst2024gpt,
  title={Gpt-4o system card},
  author={Hurst, Aaron and Lerer, Adam and Goucher, Adam P and Perelman, Adam and Ramesh, Aditya and Clark, Aidan and Ostrow, AJ and Welihinda, Akila and Hayes, Alan and Radford, Alec and others},
  journal={arXiv preprint arXiv:2410.21276},
  year={2024}
}

@article{tian2025mige,
  title={Mige: A unified framework for multimodal instruction-based image generation and editing},
  author={Tian, Xueyun and Li, Wei and Xu, Bingbing and Yuan, Yige and Wang, Yuanzhuo and Shen, Huawei},
  journal={arXiv preprint arXiv:2502.21291},
  year={2025}
}

@article{jiang2025infiniteyou,
  title={InfiniteYou: Flexible photo recrafting while preserving your identity},
  author={Jiang, Liming and Yan, Qing and Jia, Yumin and Liu, Zichuan and Kang, Hao and Lu, Xin},
  journal={arXiv preprint arXiv:2503.16418},
  year={2025}
}

@inproceedings{sun2024generative,
  title={Generative multimodal models are in-context learners},
  author={Sun, Quan and Cui, Yufeng and Zhang, Xiaosong and Zhang, Fan and Yu, Qiying and Wang, Yueze and Rao, Yongming and Liu, Jingjing and Huang, Tiejun and Wang, Xinlong},
  booktitle={Proceedings of the IEEE/CVF Conference on Computer Vision and Pattern Recognition},
  pages={14398--14409},
  year={2024}
}

@article{wang2024emu3,
  title={Emu3: Next-token prediction is all you need},
  author={Wang, Xinlong and Zhang, Xiaosong and Luo, Zhengxiong and Sun, Quan and Cui, Yufeng and Wang, Jinsheng and Zhang, Fan and Wang, Yueze and Li, Zhen and Yu, Qiying and others},
  journal={arXiv preprint arXiv:2409.18869},
  year={2024}
}

@article{gupta2022metamorph,
  title={Metamorph: Learning universal controllers with transformers},
  author={Gupta, Agrim and Fan, Linxi and Ganguli, Surya and Fei-Fei, Li},
  journal={arXiv preprint arXiv:2203.11931},
  year={2022}
}

@article{team2024chameleon,
  title={Chameleon: Mixed-modal early-fusion foundation models},
  author={Team, Chameleon},
  journal={arXiv preprint arXiv:2405.09818},
  year={2024}
}

@article{xie2024show,
  title={Show-o: One single transformer to unify multimodal understanding and generation},
  author={Xie, Jinheng and Mao, Weijia and Bai, Zechen and Zhang, David Junhao and Wang, Weihao and Lin, Kevin Qinghong and Gu, Yuchao and Chen, Zhijie and Yang, Zhenheng and Shou, Mike Zheng},
  journal={arXiv preprint arXiv:2408.12528},
  year={2024}
}

@article{zhou2024transfusion,
  title={Transfusion: Predict the next token and diffuse images with one multi-modal model},
  author={Zhou, Chunting and Yu, Lili and Babu, Arun and Tirumala, Kushal and Yasunaga, Michihiro and Shamis, Leonid and Kahn, Jacob and Ma, Xuezhe and Zettlemoyer, Luke and Levy, Omer},
  journal={arXiv preprint arXiv:2408.11039},
  year={2024}
}

@article{shi2024llamafusion,
  title={LlamaFusion: Adapting Pretrained Language Models for Multimodal Generation},
  author={Shi, Weijia and Han, Xiaochuang and Zhou, Chunting and Liang, Weixin and Lin, Xi Victoria and Zettlemoyer, Luke and Yu, Lili},
  journal={arXiv preprint arXiv:2412.15188},
  year={2024}
}

@article{pan2025metaquery,
  title={Transfer between modalities with metaqueries},
  author={Pan, Xichen and Shukla, Satya Narayan and Singh, Aashu and Zhao, Zhuokai and Mishra, Shlok Kumar and Wang, Jialiang and Xu, Zhiyang and Chen, Jiuhai and Li, Kunpeng and Juefei-Xu, Felix and others},
  journal={arXiv preprint arXiv:2504.06256},
  year={2025}
}

@article{chen2025blip3,
  title={Blip3-o: A family of fully open unified multimodal models-architecture, training and dataset},
  author={Chen, Jiuhai and Xu, Zhiyang and Pan, Xichen and Hu, Yushi and Qin, Can and Goldstein, Tom and Huang, Lifu and Zhou, Tianyi and Xie, Saining and Savarese, Silvio and others},
  journal={arXiv preprint arXiv:2505.09568},
  year={2025}
}

@article{bagel,
  title={Emerging properties in unified multimodal pretraining},
  author={Deng, Chaorui and Zhu, Deyao and Li, Kunchang and Gou, Chenhui and Li, Feng and Wang, Zeyu and Zhong, Shu and Yu, Weihao and Nie, Xiaonan and Song, Ziang and others},
  journal={arXiv preprint arXiv:2505.14683},
  year={2025}
}

@article{liao2025mogao,
  title={Mogao: An omni foundation model for interleaved multi-modal generation},
  author={Liao, Chao and Liu, Liyang and Wang, Xun and Luo, Zhengxiong and Zhang, Xinyu and Zhao, Wenliang and Wu, Jie and Li, Liang and Tian, Zhi and Huang, Weilin},
  journal={arXiv preprint arXiv:2505.05472},
  year={2025}
}

@article{qin2025lumina-image2,
  title={Lumina-image 2.0: A unified and efficient image generative framework},
  author={Qin, Qi and Zhuo, Le and Xin, Yi and Du, Ruoyi and Li, Zhen and Fu, Bin and Lu, Yiting and Yuan, Jiakang and Li, Xinyue and Liu, Dongyang and others},
  journal={arXiv preprint arXiv:2503.21758},
  year={2025}
}

@inproceedings{liu2024mmbench,
  title={Mmbench: Is your multi-modal model an all-around player?},
  author={Liu, Yuan and Duan, Haodong and Zhang, Yuanhan and Li, Bo and Zhang, Songyang and Zhao, Wangbo and Yuan, Yike and Wang, Jiaqi and He, Conghui and Liu, Ziwei and others},
  booktitle={European conference on computer vision},
  pages={216--233},
  year={2024},
  organization={Springer}
}

@inproceedings{yue2024mmmu,
  title={Mmmu: A massive multi-discipline multimodal understanding and reasoning benchmark for expert agi},
  author={Yue, Xiang and Ni, Yuansheng and Zhang, Kai and Zheng, Tianyu and Liu, Ruoqi and Zhang, Ge and Stevens, Samuel and Jiang, Dongfu and Ren, Weiming and Sun, Yuxuan and others},
  booktitle={Proceedings of the IEEE/CVF Conference on Computer Vision and Pattern Recognition},
  pages={9556--9567},
  year={2024}
}

@article{yu2023mm,
  title={Mm-vet: Evaluating large multimodal models for integrated capabilities},
  author={Yu, Weihao and Yang, Zhengyuan and Li, Linjie and Wang, Jianfeng and Lin, Kevin and Liu, Zicheng and Wang, Xinchao and Wang, Lijuan},
  journal={arXiv preprint arXiv:2308.02490},
  year={2023}
}

@article{ghosh2023geneval,
  title={Geneval: An object-focused framework for evaluating text-to-image alignment},
  author={Ghosh, Dhruba and Hajishirzi, Hannaneh and Schmidt, Ludwig},
  journal={Advances in Neural Information Processing Systems},
  volume={36},
  pages={52132--52152},
  year={2023}
}

@inproceedings{sheynin2024emu,
  title={Emu edit: Precise image editing via recognition and generation tasks},
  author={Sheynin, Shelly and Polyak, Adam and Singer, Uriel and Kirstain, Yuval and Zohar, Amit and Ashual, Oron and Parikh, Devi and Taigman, Yaniv},
  booktitle={Proceedings of the IEEE/CVF Conference on Computer Vision and Pattern Recognition},
  pages={8871--8879},
  year={2024}
}

@article{liu2025step1x,
  title={Step1x-edit: A practical framework for general image editing},
  author={Liu, Shiyu and Han, Yucheng and Xing, Peng and Yin, Fukun and Wang, Rui and Cheng, Wei and Liao, Jiaqi and Wang, Yingming and Fu, Honghao and Han, Chunrui and others},
  journal={arXiv preprint arXiv:2504.17761},
  year={2025}
}

@article{ye2025imgedit,
  title={Imgedit: A unified image editing dataset and benchmark},
  author={Ye, Yang and He, Xianyi and Li, Zongjian and Lin, Bin and Yuan, Shenghai and Yan, Zhiyuan and Hou, Bohan and Yuan, Li},
  journal={arXiv preprint arXiv:2505.20275},
  year={2025}
}

@inproceedings{xiao2025omnigen,
  title={Omnigen: Unified image generation},
  author={Xiao, Shitao and Wang, Yueze and Zhou, Junjie and Yuan, Huaying and Xing, Xingrun and Yan, Ruiran and Li, Chaofan and Wang, Shuting and Huang, Tiejun and Liu, Zheng},
  booktitle={Proceedings of the Computer Vision and Pattern Recognition Conference},
  pages={13294--13304},
  year={2025}
}

@article{li2024llavaoneversion,
  title={Llava-onevision: Easy visual task transfer},
  author={Li, Bo and Zhang, Yuanhan and Guo, Dong and Zhang, Renrui and Li, Feng and Zhang, Hao and Zhang, Kaichen and Zhang, Peiyuan and Li, Yanwei and Liu, Ziwei and others},
  journal={arXiv preprint arXiv:2408.03326},
  year={2024}
}

@article{zhao2024ultraedit,
  title={Ultraedit: Instruction-based fine-grained image editing at scale},
  author={Zhao, Haozhe and Ma, Xiaojian Shawn and Chen, Liang and Si, Shuzheng and Wu, Rujie and An, Kaikai and Yu, Peiyu and Zhang, Minjia and Li, Qing and Chang, Baobao},
  journal={Advances in Neural Information Processing Systems},
  volume={37},
  pages={3058--3093},
  year={2024}
}

@inproceedings{wei2024omniedit,
  title={Omniedit: Building image editing generalist models through specialist supervision},
  author={Wei, Cong and Xiong, Zheyang and Ren, Weiming and Du, Xeron and Zhang, Ge and Chen, Wenhu},
  booktitle={The Thirteenth International Conference on Learning Representations},
  year={2024}
}

@article{yu2024promptfix,
  title={Promptfix: You prompt and we fix the photo},
  author={Yu, Yongsheng and Zeng, Ziyun and Hua, Hang and Fu, Jianlong and Luo, Jiebo},
  journal={arXiv preprint arXiv:2405.16785},
  year={2024}
}

@article{bai2025qwen25vl,
  title={Qwen2. 5-vl technical report},
  author={Bai, Shuai and Chen, Keqin and Liu, Xuejing and Wang, Jialin and Ge, Wenbin and Song, Sibo and Dang, Kai and Wang, Peng and Wang, Shijie and Tang, Jun and others},
  journal={arXiv preprint arXiv:2502.13923},
  year={2025}
}

@article{uno,
  title={Less-to-more generalization: Unlocking more controllability by in-context generation},
  author={Wu, Shaojin and Huang, Mengqi and Wu, Wenxu and Cheng, Yufeng and Ding, Fei and He, Qian},
  journal={arXiv preprint arXiv:2504.02160},
  year={2025}
}

@article{fluxkontext,
  title={FLUX.1 Kontext: Flow Matching for In-Context Image Generation and Editing in Latent Space},
  author={Black Forest Labs},
  journal={},
  year={2025}
}

@article{tan2024ominicontrol,
  title={Ominicontrol: Minimal and universal control for diffusion transformer},
  author={Tan, Zhenxiong and Liu, Songhua and Yang, Xingyi and Xue, Qiaochu and Wang, Xinchao},
  journal={arXiv preprint arXiv:2411.15098},
  year={2024}
}

@misc{liu2024llavanext,
    title={LLaVA-NeXT: Improved reasoning, OCR, and world knowledge},
    url={https://llava-vl.github.io/blog/2024-01-30-llava-next/},
    author={Liu, Haotian and Li, Chunyuan and Li, Yuheng and Li, Bo and Zhang, Yuanhan and Shen, Sheng and Lee, Yong Jae},
    month={January},
    year={2024}
}

@misc{doubao-1.5-pro,
    author={Doubao},
    title={Doubao-1.5-pro},
    year={2025},
    howpublished={\url{https://seed.bytedance.com/zh/special/doubao_1_5_pro}},
}

@misc{FLUX,
    author={Black Forest Labs},
    title={FLUX},
    year={2024},
    howpublished={\url{https://github.com/black-forest-labs/flux}},
}

@misc{sd3-medium,
    author={Stability AI},
    title={SD3-medium},
    year={2024},
    howpublished={\url{https://stability.ai/news/stable-diffusion-3-medium}},
}

@misc{gpt4o,
    author={OpenAI},
    title={GPT-4o},
    year={2025},
    howpublished={\url{https://openai.com/index/introducing-4o-image-generation}},
}

@misc{gemini-2.0-flash,
    author={Google},
    title={Gemini 2.0 Flash},
    year={2025},
    howpublished={\url{https://developers.googleblog.com/en/experiment-with-gemini-20-flash-native-image-generation}},
}

@inproceedings{yu2025anyedit,
  title={Anyedit: Mastering unified high-quality image editing for any idea},
  author={Yu, Qifan and Chow, Wei and Yue, Zhongqi and Pan, Kaihang and Wu, Yang and Wan, Xiaoyang and Li, Juncheng and Tang, Siliang and Zhang, Hanwang and Zhuang, Yueting},
  booktitle={Proceedings of the Computer Vision and Pattern Recognition Conference},
  pages={26125--26135},
  year={2025}
}

@inproceedings{wu2025janus,
  title={Janus: Decoupling visual encoding for unified multimodal understanding and generation},
  author={Wu, Chengyue and Chen, Xiaokang and Wu, Zhiyu and Ma, Yiyang and Liu, Xingchao and Pan, Zizheng and Liu, Wen and Xie, Zhenda and Yu, Xingkai and Ruan, Chong and others},
  booktitle={Proceedings of the Computer Vision and Pattern Recognition Conference},
  pages={12966--12977},
  year={2025}
}

@article{chen2025janus,
  title={Janus-pro: Unified multimodal understanding and generation with data and model scaling},
  author={Chen, Xiaokang and Wu, Zhiyu and Liu, Xingchao and Pan, Zizheng and Liu, Wen and Xie, Zhenda and Yu, Xingkai and Ruan, Chong},
  journal={arXiv preprint arXiv:2501.17811},
  year={2025}
}

@article{lin2025uniworld,
  title={UniWorld: High-Resolution Semantic Encoders for Unified Visual Understanding and Generation},
  author={Lin, Bin and Li, Zongjian and Cheng, Xinhua and Niu, Yuwei and Ye, Yang and He, Xianyi and Yuan, Shenghai and Yu, Wangbo and Wang, Shaodong and Ge, Yunyang and others},
  journal={arXiv preprint arXiv:2506.03147},
  year={2025}
}

@article{zhang2025context,
  title={In-context edit: Enabling instructional image editing with in-context generation in large scale diffusion transformer},
  author={Zhang, Zechuan and Xie, Ji and Lu, Yu and Yang, Zongxin and Yang, Yi},
  journal={arXiv preprint arXiv:2504.20690},
  year={2025}
}

@article{zhuo2024lumina,
  title={Lumina-next: Making lumina-t2x stronger and faster with next-dit},
  author={Zhuo, Le and Du, Ruoyi and Xiao, Han and Li, Yangguang and Liu, Dongyang and Huang, Rongjie and Liu, Wenze and Zhu, Xiangyang and Wang, Fu-Yun and Ma, Zhanyu and others},
  journal={Advances in Neural Information Processing Systems},
  volume={37},
  pages={131278--131315},
  year={2024}
}

@inproceedings{qu2025tokenflow,
  title={Tokenflow: Unified image tokenizer for multimodal understanding and generation},
  author={Qu, Liao and Zhang, Huichao and Liu, Yiheng and Wang, Xu and Jiang, Yi and Gao, Yiming and Ye, Hu and Du, Daniel K and Yuan, Zehuan and Wu, Xinglong},
  booktitle={Proceedings of the Computer Vision and Pattern Recognition Conference},
  pages={2545--2555},
  year={2025}
}

@article{mao2025ace++,
  title={Ace++: Instruction-based image creation and editing via context-aware content filling},
  author={Mao, Chaojie and Zhang, Jingfeng and Pan, Yulin and Jiang, Zeyinzi and Han, Zhen and Liu, Yu and Zhou, Jingren},
  journal={arXiv preprint arXiv:2501.02487},
  year={2025}
}

@inproceedings{chen2025unireal,
  title={Unireal: Universal image generation and editing via learning real-world dynamics},
  author={Chen, Xi and Zhang, Zhifei and Zhang, He and Zhou, Yuqian and Kim, Soo Ye and Liu, Qing and Li, Yijun and Zhang, Jianming and Zhao, Nanxuan and Wang, Yilin and others},
  booktitle={Proceedings of the Computer Vision and Pattern Recognition Conference},
  pages={12501--12511},
  year={2025}
}

@inproceedings{radford2021learning,
  title={Learning transferable visual models from natural language supervision},
  author={Radford, Alec and Kim, Jong Wook and Hallacy, Chris and Ramesh, Aditya and Goh, Gabriel and Agarwal, Sandhini and Sastry, Girish and Askell, Amanda and Mishkin, Pamela and Clark, Jack and others},
  booktitle={International conference on machine learning},
  pages={8748--8763},
  year={2021},
  organization={PmLR}
}

@inproceedings{caron2021emerging,
  title={Emerging properties in self-supervised vision transformers},
  author={Caron, Mathilde and Touvron, Hugo and Misra, Ishan and J{\'e}gou, Herv{\'e} and Mairal, Julien and Bojanowski, Piotr and Joulin, Armand},
  booktitle={Proceedings of the IEEE/CVF international conference on computer vision},
  pages={9650--9660},
  year={2021}
}

@misc{ku2023viescore,
    title={VIEScore: Towards Explainable Metrics for Conditional Image Synthesis Evaluation}, 
    author={Max Ku and Dongfu Jiang and Cong Wei and Xiang Yue and Wenhu Chen},
    year={2023},
    eprint={2312.14867},
    archivePrefix={arXiv},
    primaryClass={cs.CV}
}

@article{wang2024qwen2,
  title={Qwen2-vl: Enhancing vision-language model's perception of the world at any resolution},
  author={Wang, Peng and Bai, Shuai and Tan, Sinan and Wang, Shijie and Fan, Zhihao and Bai, Jinze and Chen, Keqin and Liu, Xuejing and Wang, Jialin and Ge, Wenbin and others},
  journal={arXiv preprint arXiv:2409.12191},
  year={2024}
}

@inproceedings{Cho2024DSG,
  author    = {Jaemin Cho and Yushi Hu and Jason Baldridge and Roopal Garg and Peter Anderson and Ranjay Krishna and Mohit Bansal and Jordi Pont-Tuset and Su Wang},
  title     = {Davidsonian Scene Graph: Improving Reliability in Fine-grained Evaluation for Text-to-Image Generation},
  booktitle = {ICLR},
  year      = {2024},
}

@article{deng2024nova,
  title={Autoregressive Video Generation without Vector Quantization},
  author={Deng, Haoge and Pan, Ting and Diao, Haiwen and Luo, Zhengxiong and Cui, Yufeng and Lu, Huchuan and Shan, Shiguang and Qi, Yonggang and Wang, Xinlong},
  journal={arXiv preprint arXiv:2412.14169},
  year={2024}
}

@article{ravi2024sam,
  title={Sam 2: Segment anything in images and videos},
  author={Ravi, Nikhila and Gabeur, Valentin and Hu, Yuan-Ting and Hu, Ronghang and Ryali, Chaitanya and Ma, Tengyu and Khedr, Haitham and R{\"a}dle, Roman and Rolland, Chloe and Gustafson, Laura and others},
  journal={arXiv preprint arXiv:2408.00714},
  year={2024}
}

@article{oquab2023dinov2,
  title={Dinov2: Learning robust visual features without supervision},
  author={Oquab, Maxime and Darcet, Timoth{\'e}e and Moutakanni, Th{\'e}o and Vo, Huy and Szafraniec, Marc and Khalidov, Vasil and Fernandez, Pierre and Haziza, Daniel and Massa, Francisco and El-Nouby, Alaaeldin and others},
  journal={arXiv preprint arXiv:2304.07193},
  year={2023}
}

@article{su2024roformer,
  title={Roformer: Enhanced transformer with rotary position embedding},
  author={Su, Jianlin and Ahmed, Murtadha and Lu, Yu and Pan, Shengfeng and Bo, Wen and Liu, Yunfeng},
  journal={Neurocomputing},
  volume={568},
  pages={127063},
  year={2024},
  publisher={Elsevier}
}

@article{dao2023flashattention,
  title={Flashattention-2: Faster attention with better parallelism and work partitioning},
  author={Dao, Tri},
  journal={arXiv preprint arXiv:2307.08691},
  year={2023}
}

@article{liu2025flow,
  title={Flow-grpo: Training flow matching models via online rl},
  author={Liu, Jie and Liu, Gongye and Liang, Jiajun and Li, Yangguang and Liu, Jiaheng and Wang, Xintao and Wan, Pengfei and Zhang, Di and Ouyang, Wanli},
  journal={arXiv preprint arXiv:2505.05470},
  year={2025}
}

@article{luo2025editscore,
  title={EditScore: Unlocking Online RL for Image Editing via High-Fidelity Reward Modeling},
  author={Xin Luo and Jiahao Wang and Chenyuan Wu and Shitao Xiao and Xiyan Jiang and Defu Lian and Jiajun Zhang and Dong Liu and Zheng Liu},
  journal={arXiv preprint arXiv:2509.23909},
  year={2025}
}

@misc{wu2025qwenimagetechnicalreport,
      title={Qwen-Image Technical Report}, 
      author={Chenfei Wu and Jiahao Li and Jingren Zhou and Junyang Lin and Kaiyuan Gao and Kun Yan and Sheng-ming Yin and Shuai Bai and Xiao Xu and Yilei Chen and Yuxiang Chen and Zecheng Tang and Zekai Zhang and Zhengyi Wang and An Yang and Bowen Yu and Chen Cheng and Dayiheng Liu and Deqing Li and Hang Zhang and Hao Meng and Hu Wei and Jingyuan Ni and Kai Chen and Kuan Cao and Liang Peng and Lin Qu and Minggang Wu and Peng Wang and Shuting Yu and Tingkun Wen and Wensen Feng and Xiaoxiao Xu and Yi Wang and Yichang Zhang and Yongqiang Zhu and Yujia Wu and Yuxuan Cai and Zenan Liu},
      year={2025},
      eprint={2508.02324},
      archivePrefix={arXiv},
      primaryClass={cs.CV},
      url={https://arxiv.org/abs/2508.02324}, 
}

@misc{google2025gemini25flash,
  title        = {Introducing Gemini 2.5 Flash Image},
  author       = {{Google}},
  howpublished = {\url{https://developers.googleblog.com/en/introducing-gemini-2-5-flash-image/}},
  year         = {2025},
  note         = {Accessed: 2025-09-18}
}

@article{seedream2025seedream,
  title={Seedream 4.0: Toward Next-generation Multimodal Image Generation},
  author={Seedream, Team and Chen, Yunpeng and Gao, Yu and Gong, Lixue and Guo, Meng and Guo, Qiushan and Guo, Zhiyao and Hou, Xiaoxia and Huang, Weilin and Huang, Yixuan and others},
  journal={arXiv preprint arXiv:2509.20427},
  year={2025}
}

@article{ye2025echo,
  title={Echo-4o: Harnessing the power of gpt-4o synthetic images for improved image generation},
  author={Ye, Junyan and Jiang, Dongzhi and Wang, Zihao and Zhu, Leqi and Hu, Zhenghao and Huang, Zilong and He, Jun and Yan, Zhiyuan and Yu, Jinghua and Li, Hongsheng and others},
  journal={arXiv preprint arXiv:2508.09987},
  year={2025}
}

@article{ma2025hpsv3,
  author       = {Yuhang Ma and
                  Yunhao Shui and
                  Xiaoshi Wu and
                  Keqiang Sun and
                  Hongsheng Li},
  title        = {HPSv3: Towards Wide-Spectrum Human Preference Score},
  journal      = {CoRR},
  volume       = {abs/2508.03789},
  year         = {2025},
  url          = {https://doi.org/10.48550/arXiv.2508.03789},
  doi          = {10.48550/ARXIV.2508.03789},
  eprinttype    = {arXiv},
  eprint       = {2508.03789},
  bibsource    = {dblp computer science bibliography, https://dblp.org}
}

@misc{wang2025ovisu1technicalreport,
      title={Ovis-U1 Technical Report}, 
      author={Guo-Hua Wang and Shanshan Zhao and Xinjie Zhang and Liangfu Cao and Pengxin Zhan and Lunhao Duan and Shiyin Lu and Minghao Fu and Xiaohao Chen and Jianshan Zhao and Yang Li and Qing-Guo Chen},
      year={2025},
      eprint={2506.23044},
      archivePrefix={arXiv},
      primaryClass={cs.CV},
      url={https://arxiv.org/abs/2506.23044}, 
}

@article{wu2025qwen,
  title={Qwen-image technical report},
  author={Wu, Chenfei and Li, Jiahao and Zhou, Jingren and Lin, Junyang and Gao, Kaiyuan and Yan, Kun and Yin, Sheng-ming and Bai, Shuai and Xu, Xiao and Chen, Yilei and others},
  journal={arXiv preprint arXiv:2508.02324},
  year={2025}
}

@article{chang2025oneig,
  title={OneIG-Bench: Omni-dimensional Nuanced Evaluation for Image Generation},
  author={Chang, Jingjing and Fang, Yixiao and Xing, Peng and Wu, Shuhan and Cheng, Wei and Wang, Rui and Zeng, Xianfang and Yu, Gang and Chen, Hai-Bao},
  journal={arXiv preprint arXiv:2506.07977},
  year={2025}
}

@misc{black2024trainingdiffusionmodelsreinforcement,
      title={Training Diffusion Models with Reinforcement Learning}, 
      author={Kevin Black and Michael Janner and Yilun Du and Ilya Kostrikov and Sergey Levine},
      year={2024},
      eprint={2305.13301},
      archivePrefix={arXiv},
      primaryClass={cs.LG},
      url={https://arxiv.org/abs/2305.13301}, 
}

@article{fan2023dpok,
  title={Dpok: Reinforcement learning for fine-tuning text-to-image diffusion models},
  author={Fan, Ying and Watkins, Olivia and Du, Yuqing and Liu, Hao and Ryu, Moonkyung and Boutilier, Craig and Abbeel, Pieter and Ghavamzadeh, Mohammad and Lee, Kangwook and Lee, Kimin},
  journal={Advances in Neural Information Processing Systems},
  volume={36},
  pages={79858--79885},
  year={2023}
}

@inproceedings{chen2024enhancing,
  title={Enhancing diffusion models with text-encoder reinforcement learning},
  author={Chen, Chaofeng and Wang, Annan and Wu, Haoning and Liao, Liang and Sun, Wenxiu and Yan, Qiong and Lin, Weisi},
  booktitle={European Conference on Computer Vision},
  pages={182--198},
  year={2024},
  organization={Springer}
}

@article{cheng2025umo,
  title={UMO: Scaling Multi-Identity Consistency for Image Customization via Matching Reward},
  author={Cheng, Yufeng and Wu, Wenxu and Wu, Shaojin and Huang, Mengqi and Ding, Fei and He, Qian},
  journal={arXiv preprint arXiv:2509.06818},
  year={2025}
}

@article{mou2025dreamo,
  title={Dreamo: A unified framework for image customization},
  author={Mou, Chong and Wu, Yanze and Wu, Wenxu and Guo, Zinan and Zhang, Pengze and Cheng, Yufeng and Luo, Yiming and Ding, Fei and Zhang, Shiwen and Li, Xinghui and others},
  journal={arXiv preprint arXiv:2504.16915},
  year={2025}
}

@article{prabhudesai2023aligning,
  title={Aligning text-to-image diffusion models with reward backpropagation},
  author={Prabhudesai, Mihir and Goyal, Anirudh and Pathak, Deepak and Fragkiadaki, Katerina},
  year={2023}
}

@article{shao2024deepseekmath,
  title={Deepseekmath: Pushing the limits of mathematical reasoning in open language models},
  author={Shao, Zhihong and Wang, Peiyi and Zhu, Qihao and Xu, Runxin and Song, Junxiao and Bi, Xiao and Zhang, Haowei and Zhang, Mingchuan and Li, YK and Wu, Yang and others},
  journal={arXiv preprint arXiv:2402.03300},
  year={2024}
}

@article{xue2025dancegrpo,
  title={DanceGRPO: Unleashing GRPO on Visual Generation},
  author={Xue, Zeyue and Wu, Jie and Gao, Yu and Kong, Fangyuan and Zhu, Lingting and Chen, Mengzhao and Liu, Zhiheng and Liu, Wei and Guo, Qiushan and Huang, Weilin and others},
  journal={arXiv preprint arXiv:2505.07818},
  year={2025}
}

@misc{li2025mixgrpounlockingflowbasedgrpo,
      title={MixGRPO: Unlocking Flow-based GRPO Efficiency with Mixed ODE-SDE}, 
      author={Junzhe Li and Yutao Cui and Tao Huang and Yinping Ma and Chun Fan and Miles Yang and Zhao Zhong},
      year={2025},
      eprint={2507.21802},
      archivePrefix={arXiv},
      primaryClass={cs.AI},
      url={https://arxiv.org/abs/2507.21802}, 
}

@article{he2025tempflow,
  title={Tempflow-grpo: When timing matters for grpo in flow models},
  author={He, Xiaoxuan and Fu, Siming and Zhao, Yuke and Li, Wanli and Yang, Jian and Yin, Dacheng and Rao, Fengyun and Zhang, Bo},
  journal={arXiv preprint arXiv:2508.04324},
  year={2025}
}

@article{wang2025pref,
  title={Pref-grpo: Pairwise preference reward-based grpo for stable text-to-image reinforcement learning},
  author={Wang, Yibin and Li, Zhimin and Zang, Yuhang and Zhou, Yujie and Bu, Jiazi and Wang, Chunyu and Lu, Qinglin and Jin, Cheng and Wang, Jiaqi},
  journal={arXiv preprint arXiv:2508.20751},
  year={2025}
}

@article{li2025branchgrpo,
  title={Branchgrpo: Stable and efficient grpo with structured branching in diffusion models},
  author={Li, Yuming and Wang, Yikai and Zhu, Yuying and Zhao, Zhongyu and Lu, Ming and She, Qi and Zhang, Shanghang},
  journal={arXiv preprint arXiv:2509.06040},
  year={2025}
}

@article{huang2025competition,
  title={From Competition to Synergy: Unlocking Reinforcement Learning for Subject-Driven Image Generation},
  author={Huang, Ziwei and Shu, Ying and Fang, Hao and Long, Quanyu and Wang, Wenya and Guo, Qiushi and Ge, Tiezheng and Gan, Leilei},
  journal={arXiv preprint arXiv:2510.18263},
  year={2025}
}

@article{miao2024subject,
  title={Subject-driven text-to-image generation via preference-based reinforcement learning},
  author={Miao, Yanting and Loh, William and Kothawade, Suraj and Poupart, Pascal and Rashwan, Abdullah and Li, Yeqing},
  journal={Advances in Neural Information Processing Systems},
  volume={37},
  pages={123563--123591},
  year={2024}
}

@article{lipman2022flow,
  title={Flow matching for generative modeling},
  author={Lipman, Yaron and Chen, Ricky TQ and Ben-Hamu, Heli and Nickel, Maximilian and Le, Matt},
  journal={arXiv preprint arXiv:2210.02747},
  year={2022}
}

@article{albergo2022building,
  title={Building normalizing flows with stochastic interpolants},
  author={Albergo, Michael S and Vanden-Eijnden, Eric},
  journal={arXiv preprint arXiv:2209.15571},
  year={2022}
}

@article{dong2023dreamllm,
  title={Dreamllm: Synergistic multimodal comprehension and creation},
  author={Dong, Runpei and Han, Chunrui and Peng, Yuang and Qi, Zekun and Ge, Zheng and Yang, Jinrong and Zhao, Liang and Sun, Jianjian and Zhou, Hongyu and Wei, Haoran and others},
  journal={arXiv preprint arXiv:2309.11499},
  year={2023}
}
}

\clearpage
\setcounter{page}{1}
\maketitlesupplementary


\section{More Qualitative Results}
In this section, we present more qualitative results of OmniGen2. OmniGen2 demonstrates unified and versatile capabilities across a wide range of tasks, including text-to-image generation, image editing, in-context generation, and others. Moreover, it exhibits strong generalization abilities, allowing it to effectively tackle complex generation scenarios with remarkable consistency and fidelity.

\subsection{Text to Image}

\begin{figure*}[t]
    \setlength{\abovecaptionskip}{3pt}
    \centering
    \includegraphics[width=\linewidth]{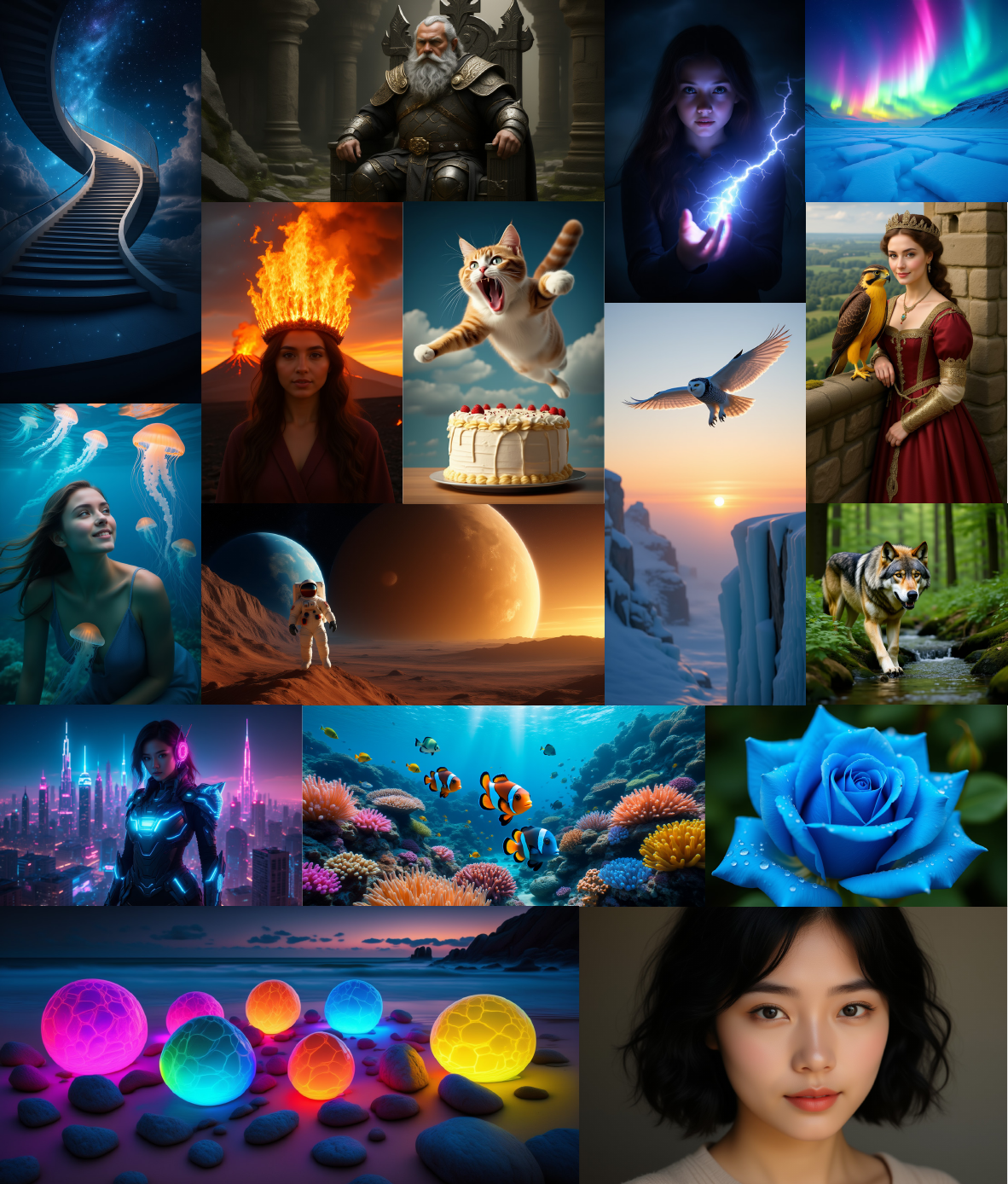}
    \caption{\textbf{Qualitative text-to-image generation by OmniGen2.} Examples showcasing the model's high fidelity to various text prompts and its support for diverse aspect ratios.}
    \label{fig:t2i_examples}
\end{figure*}

Figure~\ref{fig:t2i_examples} showcases OmniGen2's robust capabilities in text-to-image (T2I) synthesis. The model demonstrates high fidelity across a wide spectrum of conceptual and thematic prompts, from fantasy scenes like a celestial staircase to photorealistic portraits and dynamic actions. OmniGen2 excels in rendering intricate details, such as the water droplets on the blue rose, and displays a sophisticated understanding of complex lighting and composition, as seen in the dramatic glow of the girl wielding lightning and the serene ambiance of the underwater scene. Crucially, these examples also highlight the model's native support for arbitrary aspect ratios, generating high-quality portrait, landscape, and square images without distortion. This combination of conceptual diversity, high fidelity, and flexible aspect ratio support validates OmniGen2 as a powerful and versatile T2I generator.

\subsection{Image Editing}
\begin{figure*}[ht]
    \setlength{\abovecaptionskip}{3pt}
    \centering
    \includegraphics[width=\linewidth]{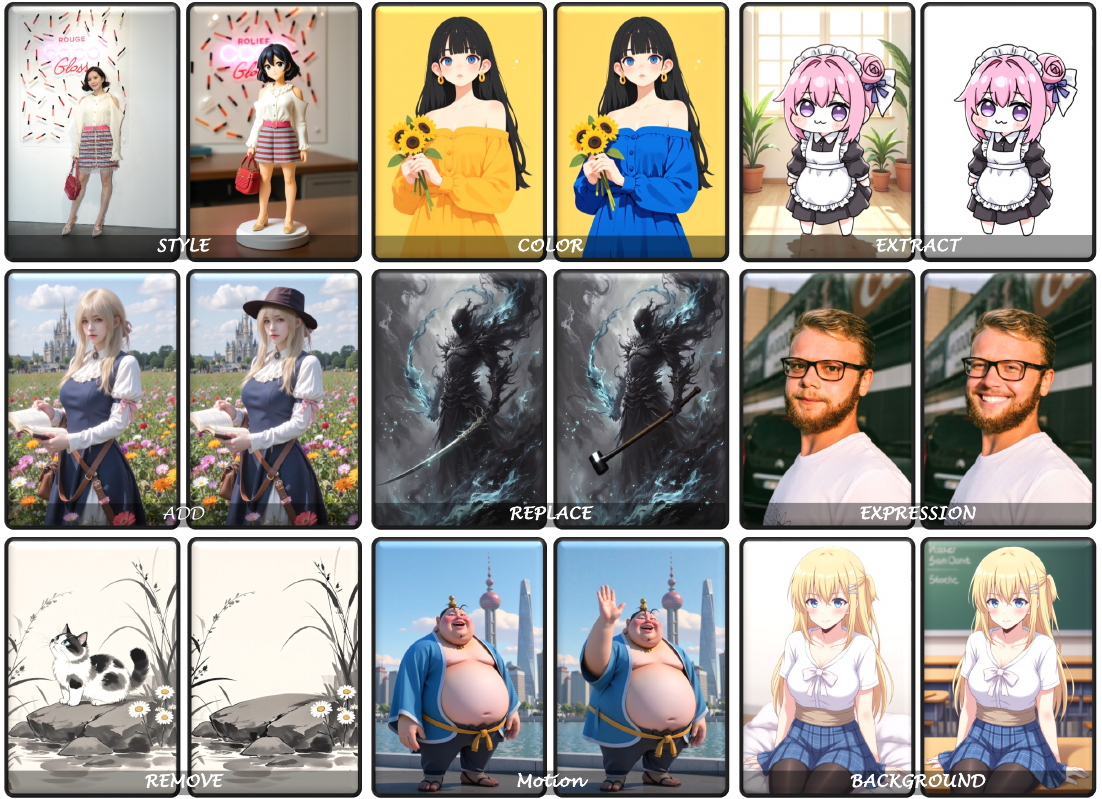}
    \caption{\textbf{Versatile image editing with OmniGen2.} The model skillfully handles a wide variety of instructions, from simple object modifications to complex motion change and stylistic alterations.}
    \label{fig:edit_examples}
    \vspace{-10pt}
\end{figure*}

Figure~\ref{fig:edit_examples} demonstrates OmniGen2's comprehensive suite of image editing capabilities, showcasing its ability to interpret a wide range of user instructions with high fidelity. The model adeptly handles localized object manipulations, including precisely adding (a hat), removing (a cat), replacing (a sword with a hammer), and extracting subjects from their backgrounds. Beyond object-level changes, OmniGen2 excels at nuanced semantic alterations, such as modifying facial expressions (adding a smile) and character motion (changing a pose to waving). Furthermore, the model is capable of executing complex, global modifications that affect the entire image, from changing backgrounds to performing complete stylistic transformations (converting a photograph into a 3D figurine). A key strength observed across all examples is the model's ability to preserve the identity of the subject and the integrity of unmodified regions, ensuring coherent and believable results.

\subsection{In context generation}
\begin{figure*}[ht]
    \setlength{\abovecaptionskip}{3pt}
    \centering
    \includegraphics[width=\linewidth]{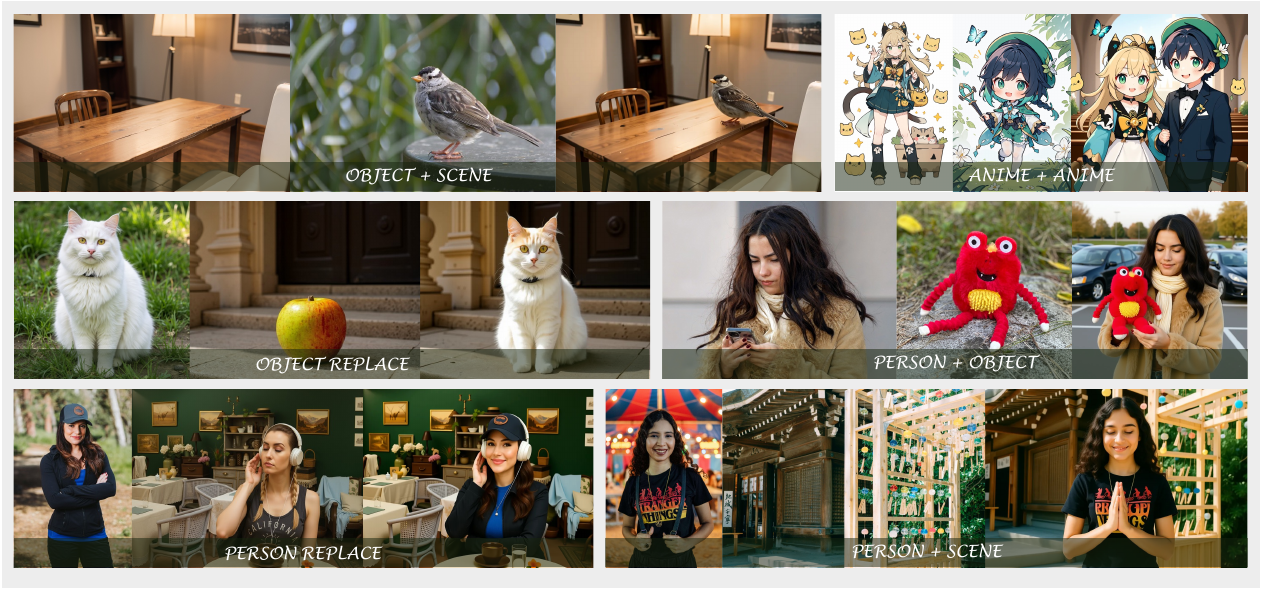}
    \caption{\textbf{Qualitative results of in-context generation and in-context edit.}}
    \label{fig:in_context_examples}
    \vspace{-10pt}
\end{figure*}

Figure~\ref{fig:in_context_examples} showcases OmniGen2's advanced capabilities in in-context generation and editing, a challenging task requiring the model to comprehend and manipulate subjects provided in reference images. The model adeptly performs compositional tasks, such as seamlessly integrating a subject into a new environment (\texttt{OBJECT + SCENE}, \texttt{PERSON + SCENE}) or combining multiple distinct subjects into a coherent new image (\texttt{ANIME + ANIME}). In these examples, OmniGen2 successfully preserves the high-fidelity identity of each subject while harmonizing them with the new context through appropriate adjustments in lighting, scale, and placement. Furthermore, the model handles complex in-context editing instructions. This includes replacing a subject within an existing scene (\texttt{OBJECT REPLACE}, \texttt{PERSON REPLACE}), where it not only swaps the main element but also intelligently adapts its appearance (e.g., color and accessories) to fit the new setting. These results demonstrate a sophisticated level of visual reasoning, where the model goes beyond simple generation to perform compositional and conditional editing based on multiple visual inputs.

\begin{figure*}[ht]
    \setlength{\abovecaptionskip}{3pt}
    \centering
    \includegraphics[width=\linewidth]{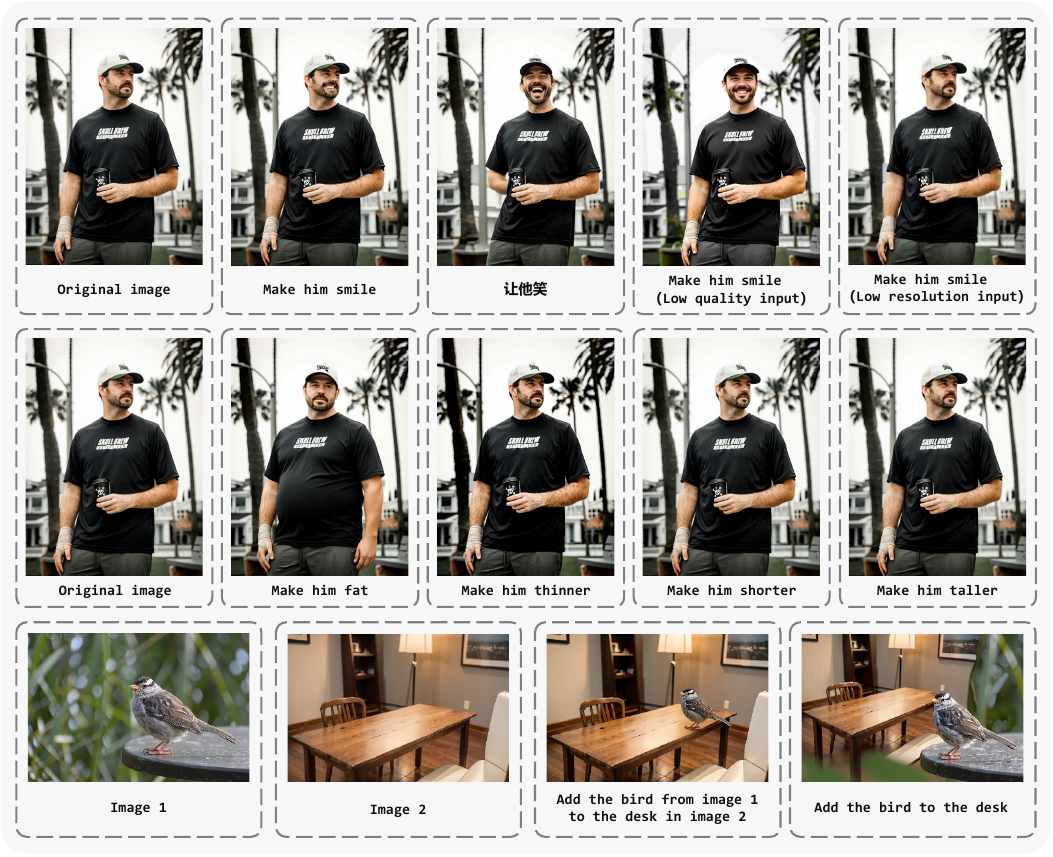}
    \caption{\textbf{Visualization of OmniGen2’s Limitations.} \textbf{Line 1:} The model performs poorly when processing Chinese prompts and low-quality images. \textbf{Line 2:} The model often struggles to modify human body shapes accurately. \textbf{Line 3:} The model is sensitive to ambiguous instructions involving multiple image sources.
}
    \label{fig:drawback}
    \vspace{-10pt}
\end{figure*}

\subsection{Limitations}

we also find several limitations of OmniGen2:

1) Performance Disparity Between English and Chinese Prompts. As shown in the first row of Figure~\ref{fig:drawback}, prompts in English generally yield better results than those in Chinese. For instance, when using a Chinese prompt, the generated image exhibits a minor inconsistency between input image and edited image.

2) Limited Generalization to Certain Instructions. The second row highlights OmniGen2’s difficulty in modifying human body shapes, likely due to the scarcity of real-world data capturing such variations.

3) Sensitivity to Input Image Quality. As illustrated in Figure~\ref{fig:drawback}, the quality of the generated output is highly sensitive to the quality of the input image. When we input a low-quality image (generated by adding noise to the raw image), the resulting images exhibit significant degradation, with details becoming notably blurred. Furthermore, downsampling the input image to a maximum dimension of 256 pixels leads to further loss of clarity and detail, and the model’s ability to accurately follow generation instructions is substantially reduced.

4) Ambiguity in Multi-Image Inputs. The third row of Figure~\ref{fig:drawback} demonstrates that the model’s performance improves when the prompt explicitly specifies the correspondence between objects and their source images (e.g., “the bird from image 1 and the desk from image 2”), indicating a sensitivity to ambiguous multi-source instructions.

5) In in-context generation tasks, the model occasionally fails to perfectly reproduce objects from the provided context. Increasing the guidance scale of image can partially alleviate this issue; however, it does not offer a complete solution. We hypothesize that significant improvements on such complex tasks may require further scaling of the model size.

\section{Other Experimental Details}

\begin{figure*}[t]
    \setlength{\abovecaptionskip}{3pt}
    \centering
    \includegraphics[width=0.65\linewidth]{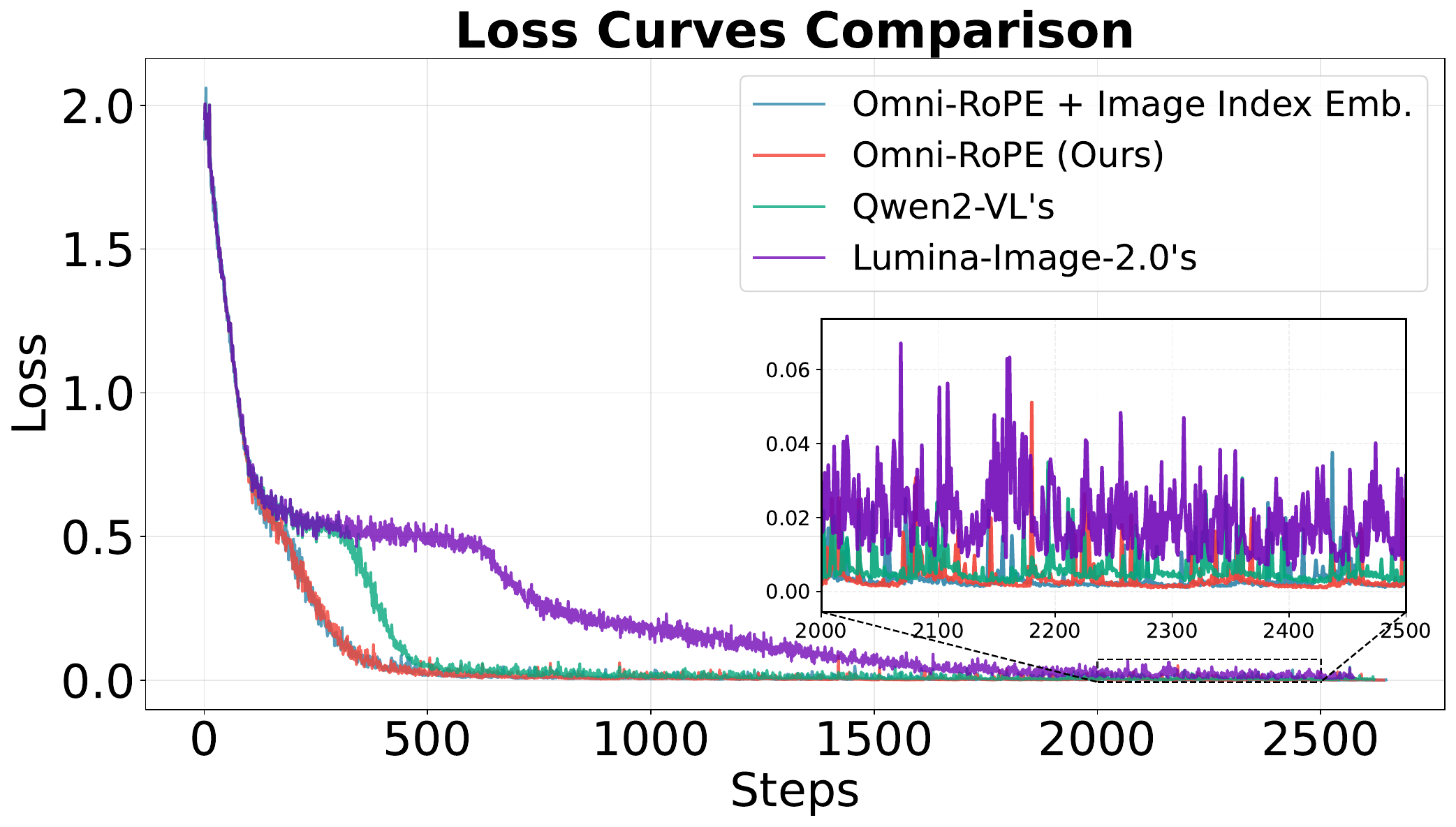}
    \caption{\textbf{Full loss curves for the Omni-RoPE toy reconstruction experiment.} Omni-RoPE converges substantially faster than prior positional encoding schemes. The inset shows late-stage optimization, where adding image index embeddings yields the lowest and most stable final loss.}
    \label{fig:rope_loss_curves}
    \vspace{-10pt}
\end{figure*}

\subsection{Toy Experiment Verification for Omni-RoPE}
Figure~\ref{fig:rope_loss_curves} provides the full loss curves for the toy reconstruction experiment introduced in Section 3.3. The trends are consistent with the quantitative results in Table 1 of the main paper: both Omni-RoPE variants reduce the reconstruction loss much faster than the prior positional encoding designs. In particular, Qwen2-VL's RoPE shows a noticeable optimization plateau before converging, while Lumina-Image-2.0's design remains unstable for a substantially longer period and converges to a worse final solution.

The zoomed-in view further highlights the late-stage behavior. Omni-RoPE already achieves a low and stable loss floor, and adding the image index embedding mainly improves the final fidelity and variance in the last stage of training, yielding the best overall reconstruction quality. These results support our key design choice of disentangling image identity from local spatial coordinates: it preserves patch-wise correspondence across images while avoiding the optimization difficulty introduced by entangled global offsets.

\subsection{Data Construction Pipeline}
For multimodal understanding tasks, we utilize the dataset provided by LLaVA-OneVision~\cite{li2024llavaoneversion}. For T2I generation, our training corpus comprises approximately 140 million open-source images sourced from Recap-DataComp~\citep{li2024recaption}, SAM-LLaVA~\citep{chen2023pixart}, ShareGPT4V~\citep{chen2023sharegpt4v}, LAION-Aesthetic~\citep{schuhmann2022laion}, ALLaVA-4V~\citep{chen2024allava}, DOCCI~\citep{onoe2024docci}, DenseFusion~\citep{li2024DenseFusion}, JourneyDB~\citep{sun2024journeydb}, and BLIP3-o~\cite{chen2025blip3}. Furthermore, we incorporate 10 million proprietary images, for which we generate synthetic annotations using the Qwen2.5-VL-72B~\cite{bai2025qwen25vl}.
For image editing tasks, we collect publicly available datasets, including SEED-Data-Edit~\citep{ge2024seededit}, UltraEdit~\cite{zhao2024ultraedit}, OmniEdit~\cite{wei2024omniedit}, PromptFix~\cite{yu2024promptfix}, and ImgEdit~\cite{ye2025imgedit}. However, these open-source resources often suffer from suboptimal image quality, limited instruction accuracy, and insufficient task diversity. To overcome these constraints and better serve our research objectives, we have meticulously constructed a new comprehensive training dataset for this study. The subsequent sections provide a detailed account of our data construction pipeline.

\subsection{In-Context Data}

The in-context image generation task~\cite{uno,ye2023ipadapter,fluxkontext,tan2024ominicontrol} focuses on extracting a visual concept—such as a specific object, identity or individual—from input images and accurately reproducing it within newly generated images. This task, also known as subject-driven generation~\cite{ruiz2023dreambooth}, parallels in-context learning in large language models: the image generation model produces personalized outputs in real time based solely on the provided context, without the need for additional fine-tuning. While in-context image generation has been extensively explored due to its broad range of applications, the community still faces a notable shortage of high-quality datasets tailored to this task.

\subsubsection{In-Context Generation}
\begin{figure*}[t]
  \centering
  \includegraphics[width=1.0\linewidth]{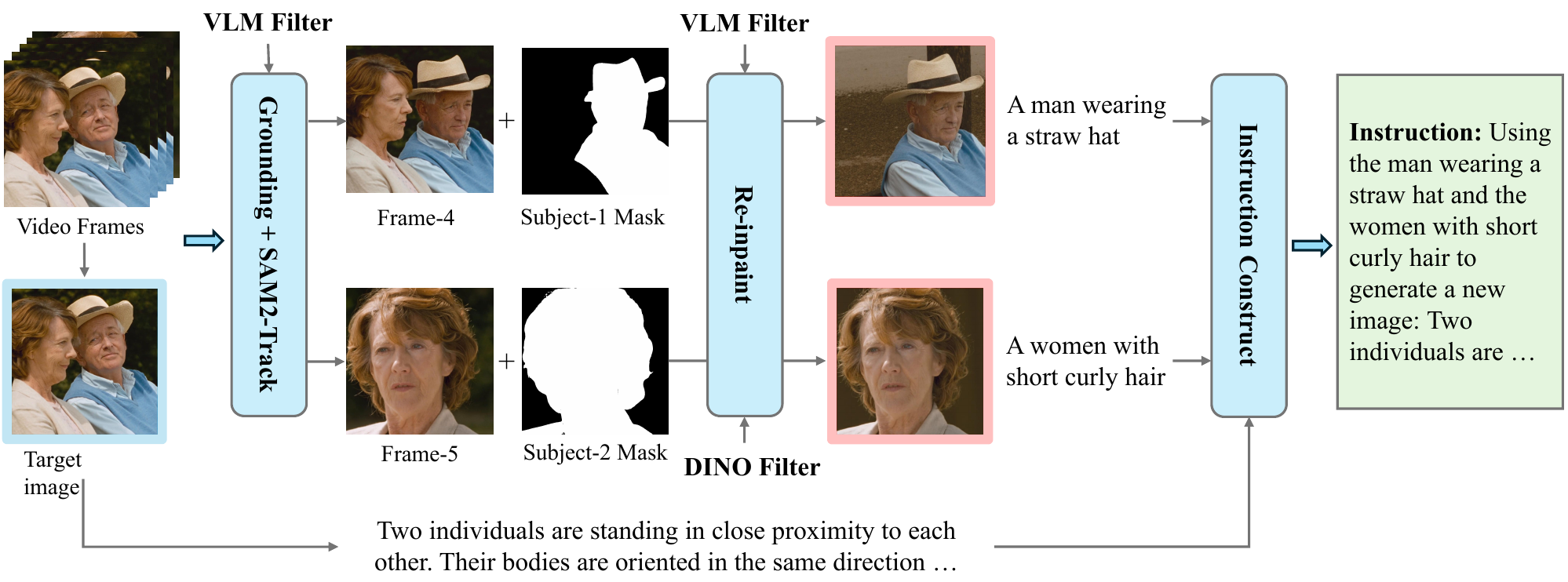}
  \caption{In-Context Generation Dataset Construction Pipeline. The final input images are outlined with a red border and the target image is marked by a blue boundary.}
  \label{fig:in-context-gen} 
  \vspace{-10pt}
\end{figure*} 

\begin{figure*}[t]
  \centering
  \includegraphics[width=1.0\linewidth]{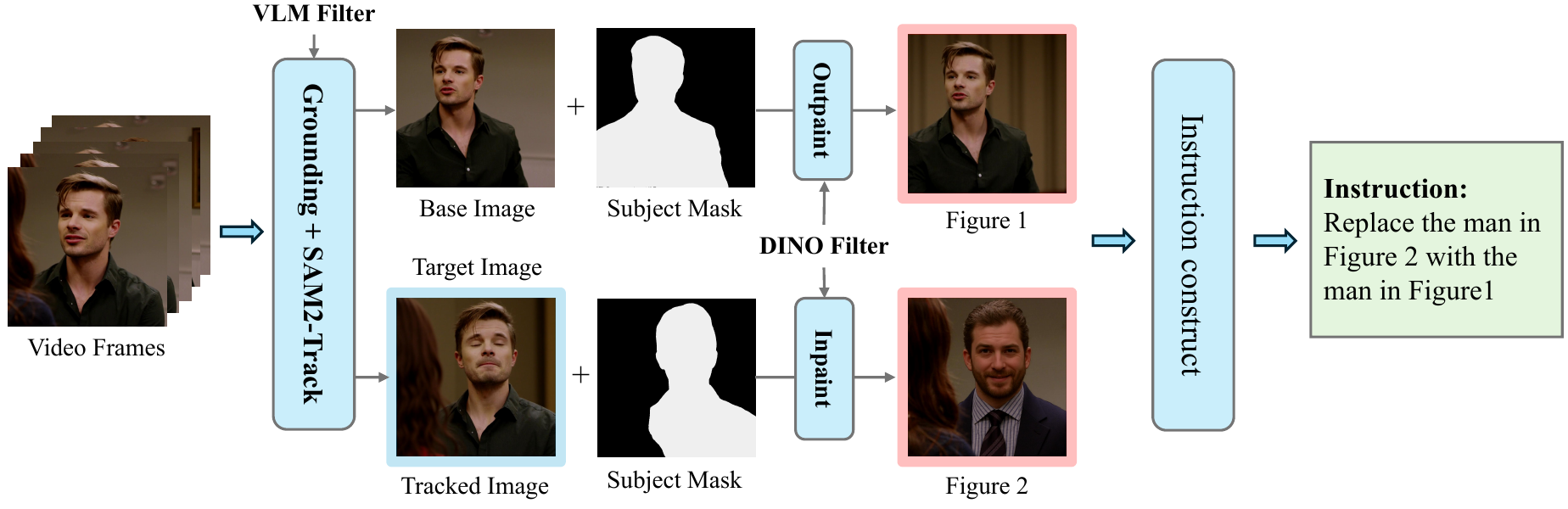}
  \vspace{-10pt}
  \caption{In-Context Editing Dataset Construction Pipeline. The final input and target images are outlined by red and blue consistent with Figure ~\ref{fig:in-context-gen}.}
  \label{fig:in-context-edit} 
  \vspace{-10pt}
\end{figure*} 

In-context generation tasks require modeling the diverse appearances of an object across different scenarios. To address this, we leverage video data, which inherently capture the same subjects under varying conditions across frames. This temporal diversity enables the construction of training pairs in which subjects remain semantically consistent but exhibit differences in pose, viewpoint, and illumination. As illustrated in Figure~\ref{fig:in-context-gen}, our data pipeline begins by extracting keyframes from each video and designating a base frame. Using Qwen2.5-VL-7B-Instruct~\cite{bai2025qwen25vl}, we identify the primary subjects within the base frame, capitalizing on the model’s vision-language capabilities to focus on semantically salient entities while filtering out irrelevant background objects. The subject bounding boxes are then obtained via Grounding DINO~\cite{groundingdino}, conditioned on the tags generated by the vision-language model. Subsequently, SAM2~\cite{ravi2024sam} is employed to segment and track identified subjects in subsequent frames, with the last valid frame containing all subjects selected to maximize appearance variation. To mitigate tracking errors—such as the inclusion of visually similar but incorrect objects—we introduce a VLM-based filtering step to ensure subject consistency. To further enhance visual diversity, FLUX.1-Fill-dev\footnote{https://huggingface.co/black-forest-labs/FLUX.1-Fill-dev} is used to outpaint the subject with a novel background in the input frame. We apply DINO~\cite{caron2021emerging}-based similarity filtering to discard samples where the subject’s appearance deviates significantly, and Qwen2.5-VL-7B-Instruct is leveraged to assess both the semantic quality and consistency of the generated samples. Additionally, Qwen2.5-VL-7B-Instruct is used to generate concise object descriptions and detailed captions for the base image, which are then integrated into natural language instructions. The final training triplet comprises the instruction, the repainted image as input, and the original image as output, providing semantically rich and visually diverse supervision for multi-subject generation tasks.
\subsubsection{In-Context Edit}

We further extend the in-context generation paradigm to editing tasks, introducing a new task termed in-context editing, as illustrated in Figure~\ref{fig:in-context-edit}. Here, the model extracts relevant elements from a context image and utilizes them to edit a target input image.

The data source for in-context editing mirrors that of in-context generation: two frames containing the same object are selected, with one serving as the context clip and the other as the target clip. Initially, object masks for both frames are obtained using SAM2~\cite{ravi2024sam}. For the context image, FLUX.1-Fill-dev is applied to generate a new background for the object via outpainting, encouraging the model to focus on object-specific features. Subsequently, FLUX.1-Fill-dev is used to inpaint the target clip, removing the object while preserving the original background to create the input clip. Finally, Qwen2.5-VL-72B-Instruct~\cite{bai2025qwen25vl} generates a natural language description of the transformation from the input clip to the target clip, which is combined with the object description from the context clip to produce comprehensive natural language instructions.

\subsection{Image Editing Data}

\begin{figure*}[t]
  \centering
  \includegraphics[width=1.0\linewidth]{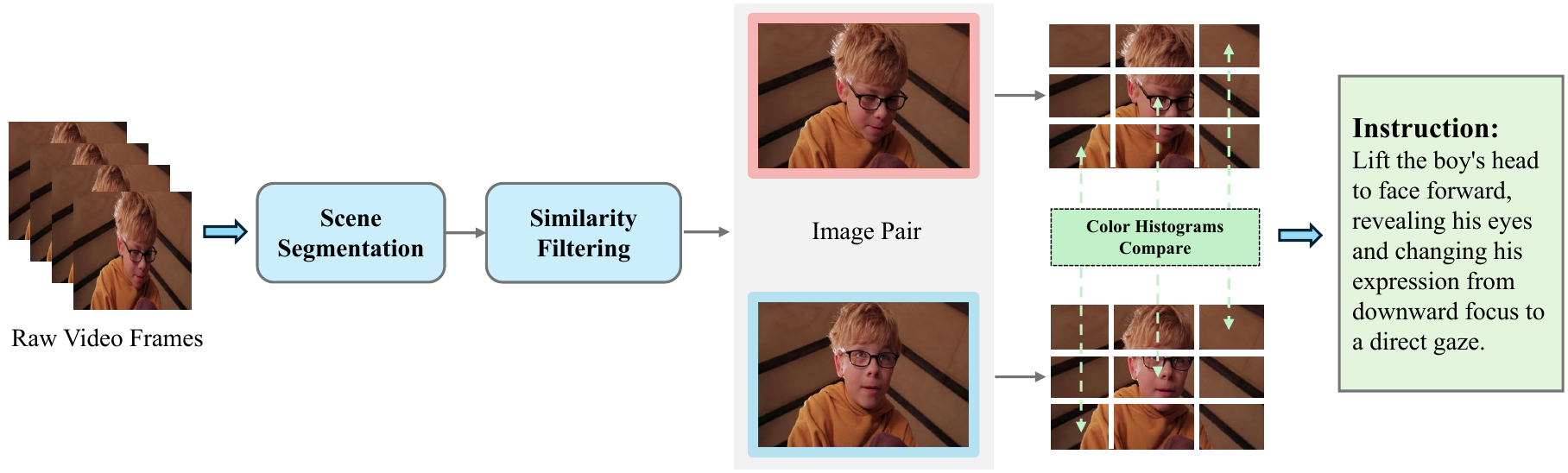}
  \vspace{-10pt}
  \caption{Create image edit pairs from videos. We first filter out frames belonging to different scenes to ensure contextual consistency, and then remove frames that exhibit significant changes in viewpoint.}
  \label{fig:video_edit} 
  \vspace{-10pt}
\end{figure*}

\subsubsection{Inpaint Data}

Although most existing editing datasets are constructed through inpainting techniques, they suffer from two primary limitations: (1) substandard image quality, stemming from both inherent low resolution and post-processing degradation during inpainting. (2) Editing instructions are inaccurate: previous work predefines editing instructions and uses inpainting models to generate images based on these instructions, but inpainting models have poor instruction-following capabilities, causing a mismatch between editing instructions and original-inpainted image pairs.

In this work, we select a small set of high-quality images from text-to-image data as our data source, applying FLUX.1-Fill-dev for inpainting. We use the inpainted images as inputs and the original images as targets to ensure high-quality target images. Additionally, we do not input instructions to the inpainting model, allowing it to fill content randomly. After obtaining image pairs, we employ a MLLM to write editing instructions based on these pairs. We find that the latest MLLM(e.g., Qwen2.5-VL) excels at writing editing instructions for original-inpainted image pairs, resulting in a high-accuracy editing dataset.

\subsubsection{Video Data}

Traditional inpainting methods are inherently limited in their capacity to construct diverse types of data, rendering them inadequate for tasks such as action modification, object movement, or expression changes. To address these limitations, we additionally extract editing pairs from video sources.

We show the pipeline in Figure~\ref{fig:video_edit}. Image editing tasks typically require localized modifications while preserving the integrity of the surrounding context. To construct suitable image editing pairs from videos, it is essential to identify frame pairs that exhibit only local changes. We begin by segmenting videos into distinct scenes to avoid pairing frames across discontinuous contexts. Scene boundaries are detected by analyzing average RGB pixel intensities, while a rolling average of differences in the HSV color space enhances robustness to rapid motion. Within each identified scene, we extract multiple frame pairs and evaluate their differences using both DINOv2~\cite{oquab2023dinov2} and CLIP~\cite{radford2021learning}. Pairs exhibiting substantial differences—indicative of viewpoint changes—or negligible differences are filtered out.

Since camera viewpoints in videos often change even within a single scene, further refinement is necessary. Existing approaches, such as vision-language models, are computationally expensive and prone to inaccuracies, while methods based on color histograms or pixel-level similarity are either insensitive to spatial structure or overly susceptible to noise. To address these challenges, we divide each image into multiple blocks and compare the color histograms of corresponding blocks to assess their similarity, effectively reducing the impact of noise. The proportion of similar blocks is then computed to impose spatial constraints, serving as a reliable indicator of viewpoint consistency. This strategy efficiently filters out frame pairs with viewpoint changes while maintaining computational efficiency.

Finally, for each retained image pair with a consistent camera viewpoint, we employ Qwen2.5-VL-72B-Instruct~\cite{bai2025qwen25vl} to generate precise editing instructions, thereby facilitating the construction of high-quality image editing datasets.

\begin{figure*}
    \centering
    \includegraphics[width=1\linewidth]{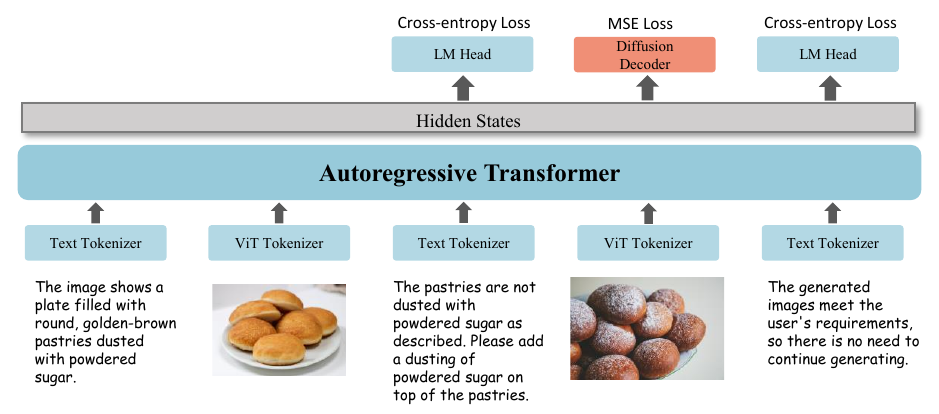}
    \vspace{-10pt}
    \caption{Multimodal Reflection for image generation.}
    \label{fig:reflection}
    \vspace{-10pt}
\end{figure*}

\begin{figure*}[!ht]
  \centering
  \includegraphics[width=1.0\linewidth]{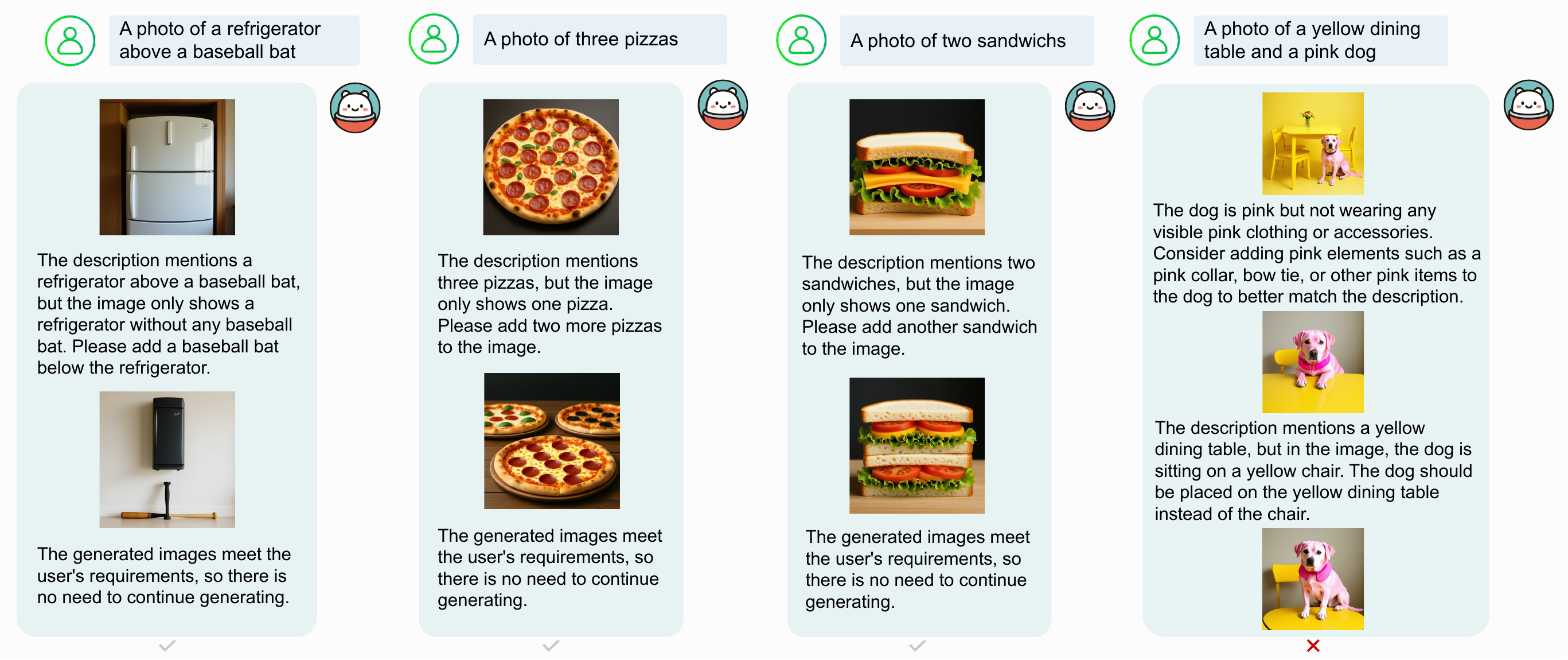}
  \caption{Example of generation with reflection using OmniGen2. \textbf{Left} and \textbf{middle:} Successful correction via one round of reflection. \textbf{Right:} an example of failed reflection, where the correct answer is incorrectly judged as wrong due to over-reflection.}
  \label{fig:reflection_example}
  \vspace{-10pt}
\end{figure*} 

\subsection{Interleave Data}
\subsubsection{Interleaved Frames}

We initially segment videos based on detected scene transitions and extract key frames from each segment. Subsequently, we construct two types of video frame sequences, each comprising up to five frames: intra-scene interleaved sequence composed of frames within identical scene and inter-scene interleaved sequence composed of frames across different scenes. Following frame sequence extraction, we annotate each pair of consecutive frames with descriptive captions using an MLLM to describe changes in object actions and behaviors, variations in environment and background, and differences in object appearances. Given the substantial volume of required annotations, we employ Qwen2.5-VL-7B-Instruct for this process. Consequently, we obtain 0.8 million interleaved data samples from video sources, which serve to pretrain the model's capacity for processing continuous multimodal sequences.

\begin{figure*}[t]
  \centering
  \includegraphics[width=0.8\linewidth]{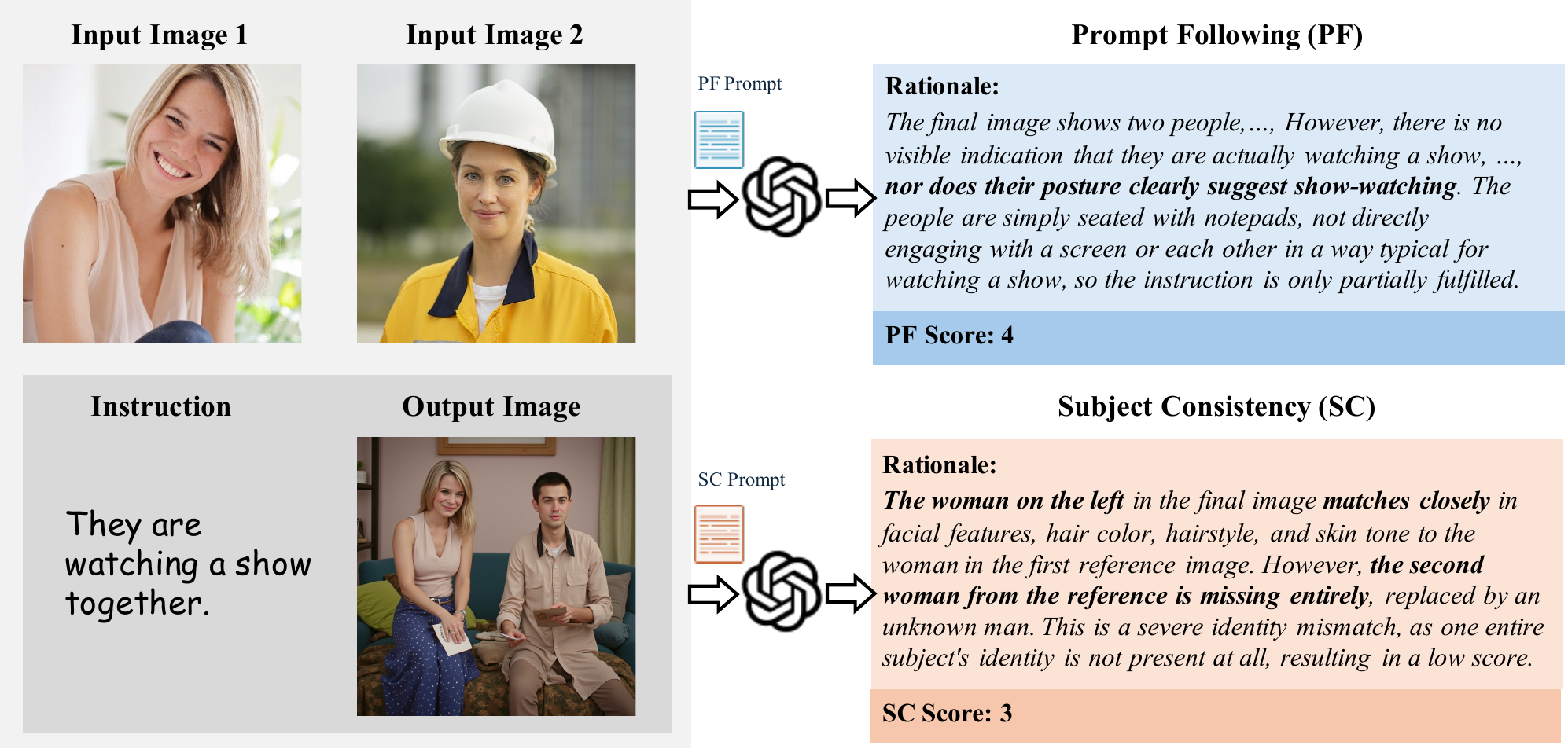}
  \caption{An illustrative example of evaluating the output image in the OmniContext benchmark.}
  \label{fig:evaluation} 
  \vspace{-12pt}
\end{figure*} 

\subsubsection{Reflection Data}
Inspired by previous advances in test-time scaling and self-reflection of large language models~\citep{guo2025deepseek, jaech2024openai, lightman2023let}, we further explore the integration of reflection capabilities into multimodal generation models and demonstrate how test-time scaling can enhance the quality of image generation. In this section, we focus on describing the construction of the reflection data for subsequent model fine-tuning. The reflection data comprise an interleaved sequence of text and images, beginning with a user instruction, followed by the multimodal model's generated image, and step-by-step reflections on the previous generated outputs. Each reflection addresses two key aspects: 1) an analysis of the deficiencies or unmet requirements in relation to the original instruction, and 2) proposed solutions to address the previous image's limitation.

To construct self-reflection data, we select a small subset from the training data (in the current experiment, we only use data from the text-to-image task) and generate images through the model. Subsequently, we use an MLLM to assess whether the generated images meet the instruction requirements. If the images fail to adequately follow instructions or exhibit other quality issues, the model identifies specific errors and suggests modifications. Initially, we experimented with the DSG~\cite{Cho2024DSG} evaluation framework to assess instruction-image alignment. However, this approach frequently led to hallucinations. Later, we discovered that powerful multimodal models could handle this task directly, so we employed Doubao-1.5-pro~\cite{doubao-1.5-pro} to output issues and modification suggestions. After obtaining the first round of reflections, we append the generated images and corresponding reflections to the original instructions and fine-tune the model on these data. Once training is complete, we continue inferring data (using the first round of reflection data) to obtain a second round of images and corresponding reflective data. This iterative process yields multiple rounds of self-reflection data.

There is currently limited research on employing reflection mechanisms to enhance image generation tasks within multimodal generative models. We hope that our present work will contribute to advancing the development of reasoning capabilities in the field of multimodal generation. After the model acquires initial reflective capabilities through training with the current data, online reinforcement learning algorithms can further enhance these capabilities, which we leave for future exploration.
\subsection{Reflection Fine-Tuning}

We fine-tune OmniGen2 on reflection dataset following the illustrated in Figure~\ref{fig:reflection}. The model's enhanced reflection capabilities are demonstrated through the examples in Figure~\ref{fig:reflection_example}. In the successful cases, the model effectively reflects on the initial generated image, identifies its shortcomings, makes appropriate corrections and terminate the generation process at an appropriate point. However, it still faces challenges in reflection and correction. The model may over-reflect on simple instructions, generating unnecessary requirements or fails to revise the image. These issues arise from the limited perception of the 3B-scale MLLM and insufficient reflection data. In future work, we plan to scale up the model and employ reinforcement learning to improve reflection quality.
\subsection{Training pipeline details}
Detailed configurations for each stage are summarized in Table~\ref{tab:training_details}.
\begin{table}[h!]
    \centering
    \resizebox{0.99\linewidth}{!}{
    \begin{tabular}{l|l|l|c}
        \toprule
        \textbf{Stage} & \textbf{Resolution} & \textbf{Task Type} & \textbf{Steps} \\
        \midrule
        \multirow{6}{*}{\textbf{1. Pre-training}} & \multirow{2}{*}{256$\times$256} & T2I-only & 50k \\
        & & Mixed-Task & 50k \\
        \cmidrule{2-4}
        \multirow{2}{*}{} & \multirow{2}{*}{512$\times$512} & T2I-only & 30k \\
        & & Mixed-Task & 30k \\
        \cmidrule{2-4}
        \multirow{2}{*}{} & \multirow{2}{*}{1024$\times$1024} & T2I-only & 50k \\
        &  & Mixed-Task & 50k \\
        \midrule
        \textbf{2. SFT} & 1024$\times$1024 & Mixed-Task & 100k \\
        \midrule
        \textbf{3. RL Alignment} & 512$\times$512 & Mixed-Task & 2.4k \\
        \bottomrule
    \end{tabular}
    }
    \caption{\textbf{Details of the OmniGen2 staged training pipeline.} The curriculum progresses from general pre-training to general instruction alignment, with increasing task complexity and resolution.}
    \label{tab:training_details}

\end{table}

\begin{table*}[!h]
    \centering
    \resizebox{0.80\linewidth}{!}{
    \begin{tabular}{l|ccc|ccc|ccc}
        \toprule
        \multirow{3}{*}{Method} & \multicolumn{9}{c}{SINGLE$\uparrow$}\\ 
        \cmidrule(lr){2-10}
        & \multicolumn{3}{c|}{Character} & \multicolumn{3}{c|}{Object} & \multicolumn{3}{c}{Average} \\
        \cmidrule(lr){2-4}
        \cmidrule(lr){5-7}        
        \cmidrule(lr){8-10}
        
        & PF & SC & Overall & PF & SC & Overall & PF & SC & Overall \\
        \midrule
        Flux.1 Kontext max~\cite{fluxkontext} & 7.98 & \textbf{9.24} & 8.48 & 8.78 & \textbf{8.76} & 8.68 & 8.38 & \textbf{9.00} & 8.58 \\
        Gemini-2.0-flash~\cite{gemini-2.0-flash} & 5.54 & 5.98 & 5.06 & 6.17 & 5.89 & 5.17 & 5.86 & 5.93 & 5.11 \\
        GPT-4o~\cite{gpt4o} & \textbf{8.89} & 9.03 & \textbf{8.90} & \textbf{9.40} & 8.74 & \textbf{9.01} & \textbf{9.14} & 8.88 & \textbf{8.95} \\
        \midrule
        InfiniteYou~\cite{jiang2025infiniteyou} & 7.81 & 5.15 & 6.05 & - & - & - & - & - & - \\
        UNO~\cite{uno} & 7.56 & 6.48 & 6.60 & 7.78 & 6.65 & 6.83 & 7.67 & 6.56 & 6.72 \\
        BAGEL~\cite{bagel} & 7.72 & 4.86 & 5.48 & 8.56 & 6.06 & 7.03 & 8.14 & 5.46 & 6.25 \\
        OmniGen~\cite{xiao2025omnigen} & 7.12 & 7.58 & 7.21 & 7.66 & 5.04 & 5.71 & 7.39 & 6.31 & 6.46 \\
        \rowcolor{myblue}\textbf{OmniGen2} & \textbf{7.92} & \textbf{8.68} & \textbf{8.19} & \textbf{8.98} & \textbf{8.44} & \textbf{8.63} & \textbf{8.45} & \textbf{8.56} & \textbf{8.41} \\
        \bottomrule
    \end{tabular}
}
\vspace{5pt}
\caption{Comparison on task type SINGLE from OmniContext. Prompt Following (PF), Subject Consistency (SC), and Overall scores are reported (higher is better, $\uparrow$).}
\label{tab:omni_context_single}
\vspace{-10pt}
\end{table*}

\begin{table*}[!h]
    \centering
    \resizebox{0.99\linewidth}{!}{
    \begin{tabular}{l|ccc|ccc|ccc|ccc}
        \toprule
        \multirow{3}{*}{Method} & \multicolumn{12}{c}{MULTIPLE$\uparrow$}\\ 
        \cmidrule(lr){2-13}
        & \multicolumn{3}{c|}{Character} & \multicolumn{3}{c|}{Object} & \multicolumn{3}{c|}{Char. + Obj.} & \multicolumn{3}{c}{Average} \\
        \cmidrule(lr){2-4}
        \cmidrule(lr){5-7}        
        \cmidrule(lr){8-10}
        \cmidrule(lr){11-13}
        & PF & SC & Overall & PF & SC & Overall & PF & SC & Overall & PF & SC & Overall \\
        \midrule
        Gemini-2.0-flash~\cite{gemini-2.0-flash} & 3.65 & 3.02 & 2.91 & 2.50 & 5.02 & 2.16 & 4.26 & 5.80 & 3.80 & 3.47 & 4.62 & 2.96 \\
        GPT-4o~\cite{gpt4o} & \textbf{9.17} & \textbf{9.03} & \textbf{9.07} & \textbf{9.06} & \textbf{8.90} & \textbf{8.95} & \textbf{8.34} & \textbf{8.89} & \textbf{8.54} & \textbf{8.86} & \textbf{8.94} & \textbf{8.86} \\
        \midrule
        UNO~\cite{uno} & 3.88 & 2.38 & 2.54 & 7.46 & 5.86 & 6.51 & 5.10 & 4.10 & 4.39 & 5.48 & 4.11 & 4.48 \\
        BAGEL~\cite{bagel} & 6.14 & 4.86 & 5.17 & 7.54 & 6.10 & 6.64 & 6.74 & 6.28 & 6.24 & 6.81 & 5.75 & 6.02 \\
        OmniGen~\cite{xiao2025omnigen} & 5.92 & 6.18 & 5.65 & 5.60 & 5.46 & 5.44 & 4.64 & 4.96 & 4.68 & 5.39 & 5.53 & 5.26\\
        \rowcolor{myblue}\textbf{OmniGen2} & \textbf{7.30} & \textbf{8.10} & \textbf{7.45} & \textbf{7.98} & \textbf{7.74} & \textbf{7.80} & \textbf{7.60} & \textbf{8.34} & \textbf{7.93} & \textbf{7.63} & \textbf{8.06} & \textbf{7.73} \\
        \bottomrule
    \end{tabular}
}
\vspace{5pt}
    \caption{Comparison on task type MULTIPLE from OmniContext. Prompt Following (PF), Subject Consistency (SC), and Overall scores are reported (higher is better, $\uparrow$).}
\label{tab:omni_context_multiple}
\vspace{-10pt}
\end{table*}

\begin{table*}[!h]
    \centering
    \resizebox{0.99\linewidth}{!}{
    \begin{tabular}{l|ccc|ccc|ccc|ccc}
        \toprule
        \multirow{3}{*}{Method} & \multicolumn{12}{c}{SCENE$\uparrow$}\\ 
        \cmidrule(lr){2-13}
        & \multicolumn{3}{c|}{Character} & \multicolumn{3}{c|}{Object} & \multicolumn{3}{c|}{Char. + Obj.} & \multicolumn{3}{c}{Average} \\
        \cmidrule(lr){2-4}
        \cmidrule(lr){5-7}        
        \cmidrule(lr){8-10}
        \cmidrule(lr){11-13}
        & PF & SC & Overall & PF & SC & Overall & PF & SC & Overall & PF & SC & Overall \\
        \midrule
        Gemini-2.0-flash~\cite{gemini-2.0-flash} & 3.76 & 3.33 & 3.02 & 4.02 & 5.22 & 3.89 & 2.89 & 4.63 & 2.92 & 3.56 & 4.39 & 3.28 \\
        GPT-4o~\cite{gpt4o} & \textbf{9.05} & \textbf{8.88} & \textbf{8.90} & \textbf{8.33} & \textbf{8.62} & \textbf{8.44} & \textbf{8.71} & \textbf{8.57} & \textbf{8.60} & \textbf{8.70} & \textbf{8.69} & \textbf{8.65} \\
        \midrule
        UNO~\cite{uno} & 2.74 & 2.50 & 2.06 & 5.62 & 3.52 & 4.33 & 5.22 & 3.86 & 4.37 & 4.53 & 3.29 & 3.59 \\
        BAGEL~\cite{bagel} & 4.56 & 3.94 & 4.07 & 6.12 & 5.50 & 5.71 & 5.90 & 5.30 & 5.47 & 5.53 & 4.91 & 5.08 \\
        OmniGen~\cite{xiao2025omnigen} & 4.14 & 3.42 & 3.59 & 5.24 & 3.72 & 4.32 & 5.56 & 4.84 & 5.12 & 4.98 & 3.99 & 4.34 \\
        \rowcolor{myblue}\textbf{OmniGen2} & \textbf{8.02} & \textbf{7.64} & \textbf{7.75} & \textbf{8.10} & \textbf{7.80} & \textbf{7.91} & \textbf{8.08} & \textbf{7.88} & \textbf{7.93} & \textbf{8.07} & \textbf{7.77} & \textbf{7.86} \\
        \bottomrule
    \end{tabular}
}
\vspace{5pt}
\caption{Comparison on task type SCENE from OmniContext. Prompt Following (PF), Subject Consistency (SC), and Overall scores are reported (higher is better, $\uparrow$).}
\label{tab:omni_context_scene}
\vspace{-10pt}
\end{table*}

\subsection{OmniContext Details}
The detailed metrics for each subtask are presented in Tables~\ref{tab:omni_context_single},~\ref{tab:omni_context_multiple} and ~\ref{tab:omni_context_scene}. The pipeline to evaluate models on OmniContext as shown in Figure~\ref{fig:evaluation}.

\subsection{RL Generalization to Out-of-Distribution Benchmarks}
To further evaluate the generalization ability of our multi-stage RL curriculum, we conduct experiments on out-of-distribution (OOD) benchmarks that are not directly aligned with our training rewards.

Specifically, we evaluate on Emu-Edit and OneIG-Bench, which assess editing fidelity and general image generation quality under diverse and challenging conditions. These benchmarks differ from our training objectives and thus provide a reliable measure of transferability.

As shown in Table~\ref{tab:ood_ablation}, our full curriculum (Edit → GenEval → IC) consistently outperforms the base model and alternative training orders across all OOD metrics. This demonstrates that our alignment strategy does not overfit to specific reward signals, but instead learns generalizable capabilities that transfer across tasks.
\begin{table}[t]
\centering
\footnotesize
\setlength{\tabcolsep}{2.5pt}
\begin{tabular}{lcccc}
\toprule
\multirow{2}{*}{Strategy} & \multicolumn{3}{c}{Emu-Edit} & OneIG \\
\cmidrule(lr){2-4} \cmidrule(lr){5-5}
 & CLIP-I$\uparrow$ & CLIP-Out$\uparrow$ & DINO$\uparrow$ & Align.$\uparrow$ \\
\midrule
Base & 0.877 & 0.309 & 0.823 & 0.7870 \\
GenEval+Edit & 0.909 & 0.311 & 0.894 & 0.8160 \\
GenEval+Edit+IC & \underline{0.886} & \textbf{0.312} & \underline{0.858} & 0.8212 \\
Edit+IC+GenEval & 0.868 & 0.310 & 0.826 & \underline{0.8242} \\
\textbf{Edit+GenEval+IC} & \textbf{0.896} & \underline{0.311} & \textbf{0.876} & \textbf{0.8289} \\
\bottomrule
\end{tabular}
\caption{RL curriculum ablation on out-of-distribution (OOD) benchmarks. Base denotes the model without RL.}
\label{tab:ood_ablation}
\end{table}

\subsection{GenEval Results}
As shown in the Table~\ref{tab:geneval}, OmniGen2 excels at generating images from complex,
compositional prompts. Our model achieves an impressive overall score of 0.95. This result surpasses other powerful unified models like UniWorld-V1 (0.84) and BAGEL (0.88). It is crucial to note that this SOTA performance is achieved with exceptional efficiency. OmniGen2 utilizes only 4B trainable parameters and was trained on 15M T2I pairs and 50k prompts used in RL. 

\begin{table}[t]
    \centering
    \resizebox{0.99\linewidth}{!}{
        \begin{tabular}{lcccccccc}
            \toprule
            Method & \multicolumn{1}{c}{Single object$\uparrow$} & \multicolumn{1}{c}{Two object$\uparrow$} & \multicolumn{1}{c}{Counting$\uparrow$} & \multicolumn{1}{c}{Colors$\uparrow$} & \multicolumn{1}{c}{Position$\uparrow$} & \multicolumn{1}{c}{Color attribution$\uparrow$} & \multicolumn{1}{c}{Overall$\uparrow$} \\
            \midrule
            SDv2.1~\cite{rombach2022high} & 0.98 & 0.51 & 0.44 & 0.85 & 0.07 & 0.17 & 0.50 \\
            SDXL~\cite{podell2023sdxl} & 0.98 & 0.74 & 0.39 & 0.85 & 0.15 & 0.23 & 0.55 \\
            IF-XL & 0.97 & 0.74 & 0.66 & 0.81 & 0.13 & 0.35 & 0.61 \\
            LUMINA-Next~\cite{zhuo2024lumina} & 0.92 & 0.46 & 0.48 & 0.70 & 0.09 & 0.13 & 0.46 \\
            SD3-medium~\cite{sd3-medium} & 0.99 & 0.94 & 0.72 & 0.89 & 0.33 & 0.60 & 0.74 \\
            FLUX.1-dev~\cite{FLUX} & 0.99 & 0.81 & 0.79 & 0.74 & 0.20 & 0.47 & 0.67 \\
            NOVA~\cite{deng2024nova} & 0.99 & 0.91 & 0.62 & 0.85 & 0.33 & 0.56 & 0.71 \\
            OmniGen~\cite{xiao2025omnigen} & 0.98 & 0.84 & 0.66 & 0.74 & 0.40 & 0.43 & 0.68 \\ 
            \midrule
            TokenFlow-XL~\cite{qu2025tokenflow} & 0.95 & 0.60 & 0.41 & 0.81 & 0.16 & 0.24 & 0.55 \\ 
            Janus~\cite{wu2025janus} & 0.97 & 0.68 & 0.30 & 0.84 & 0.46 & 0.42 & 0.61 \\
            Janus Pro~\cite{chen2025janus} & 0.99 & 0.89 & 0.59 & 0.90 & 0.79 & 0.66 & 0.80 \\
            $\text{Emu3-Gen}^{\dagger}$~\cite{wang2024emu3} & 0.99 & 0.81 & 0.42 & 0.80 & 0.49 & 0.45 & 0.66 \\
            Show-o~\cite{xie2024show} & 0.98 & 0.80 & 0.66 & 0.84 & 0.31 & 0.50 & 0.68 \\
            $\text{MetaQuery-XL}^{\dagger}$~\cite{pan2025metaquery} & - & - & - & - & - & - & 0.80 \\
            $\text{BLIP3-o}^{\dagger}$ 4B~\cite{chen2025blip3} & - & - & - & - & - & - & 0.81 \\
            $\text{BLIP3-o}^{\dagger}$ 8B~\cite{chen2025blip3} & - & - & - & - & - & - & 0.84 \\
            BAGEL~\cite{bagel} & 0.99 & 0.94 & 0.81 & 0.88 & 0.64 & 0.63 & 0.82 \\
            $\text{BAGEL}^{\dagger}$~\cite{bagel} & 0.98 & 0.95 & 0.84 & 0.95 & 0.78 & 0.77 & 0.88 \\
            UniWorld-V1~\cite{lin2025uniworld} & 0.99 & 0.93 & 0.79 & 0.89 & 0.49 & 0.70 & 0.80 \\
            $\text{UniWorld-V1}^{\dagger}$~\cite{lin2025uniworld} & 0.98 & 0.93 & 0.81 & 0.89 & 0.74 & 0.71 & 0.84 \\
            \rowcolor{myblue}\textbf{OmniGen2} & 0.99 & 1 & 0.93 & 0.91 & 1 & 0.86 & \textbf{0.95} \\ 
            
            \bottomrule
        \end{tabular}
    }
    \vspace{5pt}
    \caption{Evaluation of text-to-image generation ability on GenEval~\cite{ghosh2024geneval} benchmark. $\dagger$ refers to the methods using LLM rewriter.}
    \label{tab:geneval}
    	\vspace{-5pt}
\end{table}

\end{document}